\documentclass[final,5p,times,twocolumn]{elsarticle}

\usepackage{amsmath,amssymb,amsfonts}
\usepackage{graphicx}
\usepackage{booktabs}
\usepackage{tabularx}
\usepackage{array}
\usepackage{xcolor}
\newcolumntype{Y}{>{\centering\arraybackslash}X}
\graphicspath{{figures/}}

\newcommand{\revision}[1]{#1}
\newenvironment{revisionblock}{}{}

\usepackage[ruled,linesnumbered]{algorithm2e}
\SetAlgoLined
\SetKwInput{KwIn}{Input}
\SetKwInput{KwOut}{Output}
\usepackage[hidelinks]{hyperref}

\begin{document}

\begin{frontmatter}

\title{Target-Guided Selective Reweighting for Physics-Informed Neural Network Inverse Problems: A Transfer Learning Approach}

\author[1]{Qian Hu}

\author[1]{Bin Fan\corref{cor1}}
\ead{bfan@fjut.edu.cn}

\author[1]{Yao Xiao}

\author[1]{Zhicheng Lin}

\author[1]{Meixin Xiong}

\affiliation[1]{organization={School of Computing and Data Science, Fujian University of Technology},
            city={Fuzhou},
            postcode={350118},
            state={Fujian},
            country={China}}
\cortext[cor1]{Corresponding author}

\begin{abstract}

\begin{revisionblock}
Physics-informed neural networks (PINNs) often face ill-posed optimization, competing losses, and parameter compensation in partial differential equation (PDE) inverse problems. Transfer learning can reuse source-task representations, but direct fine-tuning may induce negative transfer when source and target physics differ, leading to low field error but inaccurate parameter recovery. To address this issue, we propose Target-Guided Selective Reweighting PINN (TGSR-PINN), a target-evidence-driven representation correction method for PINN inverse transfer learning. TGSR-PINN transfers source network weights and biases but initializes target physical parameters independently. After target short adaptation, it scores neurons using first-order Taylor sensitivity and pre-activation variance on fixed batches. These scores are converted into continuous weak-adaptation signals using a Gaussian mixture model with rank fallback. TGSR-PINN then applies bounded selective soft decay to the corresponding input weight rows and biases without pruning or resetting them. Experiments on a zero-source high-P\'{e}clet inflow--outflow problem with nonzero Dirichlet data and an outflow boundary layer, Allen--Cahn to Burgers cross-PDE transfer, and 5\%-noise reaction--diffusion inverse problems show that TGSR-PINN improves parameter recovery while maintaining low field error. Ablation studies indicate that neuron target scoring, weak-adaptation estimation, layer protection, and selective soft decay jointly contribute to the observed benefits.
\end{revisionblock}

\end{abstract}


\end{frontmatter}

\section{Introduction}
\label{sec:introduction}

Partial differential equations (PDEs) are fundamental mathematical tools for describing continuous physical systems, widely used in fluid mechanics, heat and mass transfer, elasticity, electromagnetic fields, reaction--diffusion systems, and biomedical engineering. Classical numerical methods, such as finite difference, finite element, and finite volume methods, have established mature theoretical and engineering frameworks for solving PDE forward problems. These methods typically require known governing equations, boundary conditions, and physical parameters, and solve for the target physical field through mesh discretization. However, in many practical engineering scenarios, researchers face not only forward problems but also inverse problems that require inferring unknown material properties, diffusion coefficients, velocity parameters, source terms, or boundary conditions from limited, sparse, or noisy observation data. Inverse problems are typically ill-posed and exhibit parameter correlations, where small errors in observational data may lead to significant fluctuations in inversion results, and different parameter combinations may produce similar physical field responses. Consequently, PDE inverse problems are not merely numerical solution tasks but also parameter estimation problems influenced by observation noise, physical priors, and optimization procedures.

Physics-informed neural networks provide a unified differentiable modeling framework for both forward and inverse PDE problems~\cite{raissi2019,karniadakis2021}. PINNs approximate the unknown physical field using a neural network $u_{\theta}(x,t)$ and compute PDE residuals through automatic differentiation, incorporating governing equations, initial conditions, boundary conditions, and observational data into the loss function. For inverse problems, unknown physical parameters can serve as trainable variables, optimized simultaneously with network weights and biases, thereby completing field reconstruction and parameter inversion within the same framework. Compared with purely data-driven models, PINNs leverage physical constraints to reduce reliance on large-scale labeled data and are applicable to scenarios with sparse or incomplete observations; however, the comparative advantages of PINNs over traditional methods such as finite elements vary in terms of accuracy, computational cost, and problem settings~\cite{grossmann2024}. Subsequent research has further integrated PINNs into physics-informed machine learning and scientific machine learning frameworks, advancing their application to complex engineering problems~\cite{karniadakis2021,cuomo2022,guo2025}. Recent surveys have also summarized the development and challenges of PINNs from perspectives including loss function design, geometric modeling, and structural engineering applications~\cite{plankovskyy2025,baniya2026}.

\begin{revisionblock}
PINN training is not inherently stable despite its unified formulation. A typical PINN loss combines PDE residuals, boundary conditions, initial conditions, and observational errors. These terms can differ substantially in scale, gradient direction, and convergence rate.
\end{revisionblock}
Existing studies have analyzed PINN training difficulties and applicability conditions from the perspectives of gradient pathology, neural tangent kernel, and theoretical convergence~\cite{wang2021gradient,wang2022ntk,shin2020}, and further identified that training failure modes and complex loss landscapes increase optimization difficulty~\cite{krishnapriyan2021,rathore2024}. In inverse problems, these difficulties are further amplified. On the one hand, observational data are often limited and noisy; on the other hand, the parameters to be inverted may suffer from insufficient identifiability or mutual compensation. Even when a model can reduce PDE residuals or field prediction errors, it may not accurately recover the target physical parameters. Particularly in multi-parameter inversion tasks, the network representation may compensate for incorrect parameters by adjusting field predictions, yielding seemingly reasonable field results while physical parameters remain far from their true values.

To address PINN training difficulties, numerous improvement directions have been proposed, including gradient enhancement and adaptive loss weighting~\cite{yu2022,mcclenny2023,xiang2022}, domain decomposition~\cite{jagtap2020,moseley2023}, frequency-domain modeling and augmented Lagrangian training~\cite{musgrave2024,son2023}, and residual decay or loss-attention-based weight balancing methods~\cite{chen2025,song2024}. These methods primarily focus on improving convergence stability and prediction accuracy on a single task. In contrast, PINN transfer learning, especially for inverse problems, \revision{also requires evaluating how source representations affect target parameter recovery}. For inverse problems, the value of transfer cannot be judged solely by training speed or field error; it must also be assessed by whether the transferred representation affects the quality of target physical parameter recovery.

The fundamental idea of transfer learning is to leverage knowledge learned from a source task to assist target task training. When source and target tasks share partial physical structures, geometric features, or solution-space patterns, the network weights and hidden representations from the source model can serve as an initialization for the target task, reducing the cost of training from scratch and potentially improving convergence speed and training stability~\cite{pan2010,lin2024,zhou2024}. Transfer learning has been applied to PINN inverse problems and data-guided inverse problems~\cite{lin2024,zhou2024}, and extended to structural mechanics simulation, vortex-induced vibration, and complex engineering systems~\cite{kapoor2024,tang2022}; recent work has also compared full fine-tuning and lightweight fine-tuning PINN transfer strategies~\cite{wang2025transfer}. These studies demonstrate that when source and target tasks exhibit strong relevance, the source model can provide an effective starting point for target training.

However, in PINN inverse transfer learning, source model representations are not uniformly beneficial. The source model may contain physical structures useful for the target task, but may also carry source-task biases inconsistent with the target physics. When source and target tasks differ in governing parameters, boundary conditions, observation layouts, noise levels, or dominant physical mechanisms, directly reusing the source model may steer target optimization toward regions detrimental to parameter recovery, thereby creating negative transfer risk~\cite{liu2023,yosinski2014}. This risk may not be directly observable from field errors. The source model may accelerate target loss reduction and even achieve low field prediction errors, but simultaneously affect parameter recovery through compensation between network representations and physical parameters. Engineering inverse problems such as material characterization, nondestructive testing, and structural load identification all demonstrate that physical parameter recovery is often as important as field reconstruction, or even a more direct task objective~\cite{lee2023,shukla2020}.

\begin{revisionblock}
Existing PINN transfer strategies usually operate at the layer or parameter-block level. Full fine-tuning updates the transferred network, frozen-layer methods update selected layers, and partial transfer copies a predefined subset. Because these choices are made before target training, they do not reassess individual neurons after exposure to target data. This limitation matters in inverse problems: the network representation may compensate for inaccurate physical parameters, allowing field and parameter errors to diverge. We therefore treat PINN inverse transfer as target-conditioned representation adaptation and evaluate the target support received by each transferred hidden neuron.
\end{revisionblock}

This problem is related to neuron importance estimation and pruning research. Taylor saliency and related methods approximate neuron importance through first-order or second-order loss changes and are widely used for model pruning and compression~\cite{molchanov2019,han2015learning,han2016deep}; selective pruning has also been applied to weaken noise-affected representations in PINN inverse problems~\cite{chen2026unlearning}. \begin{revisionblock}Our setting has a different objective. Low-scoring neurons may still contain reusable source information. Hard pruning, sparse subnetwork search such as the lottery ticket hypothesis~\cite{frankle2019}, or random reinitialization can remove or disrupt that information. TGSR-PINN instead attenuates these neurons progressively according to target-side evidence.\end{revisionblock}

\begin{revisionblock}
Based on these considerations, we propose Target-Guided Selective Reweighting PINN (TGSR-PINN). Only source network weights and biases are transferred; target physical parameters are initialized independently and optimized under the target loss. After target short adaptation, the method scores hidden neurons using Taylor sensitivity and pre-activation variance over fixed batches. A conditional GMM with rank fallback converts the scores into weak-adaptation signals, and layer protection with a continuous mapping determines the decay factors. Selective soft decay preserves the network topology and keeps all parameters trainable. The resulting sequence is neuron target score $\to$ weak-adaptation signal $\to$ low-scoring neuron $\to$ selective soft decay. We evaluate both field and physical-parameter errors in inverse-transfer settings with substantial source--target differences, strong parameter coupling, and possible field--parameter error decoupling.
\end{revisionblock}

The main contributions of this paper are as follows:
\begin{enumerate}
\item A target-evidence-driven representation correction framework is proposed for PINN inverse transfer learning. Unlike directly transferring the complete source-task state, TGSR-PINN explicitly transfers only network weights and biases, while target physical parameters are independently initialized by the target task and updated under the target loss. This design focuses the investigation on the impact of transferred representations on parameter recovery.
\item A neuron-level target scoring mechanism is designed. This mechanism combines first-order Taylor sensitivity and pre-activation variance under the target loss to compute empirical neuron target scores for hidden neurons on fixed scoring batches, providing target-side evidence for subsequent weak-adaptation signal estimation.
\item A weak-adaptation signal estimation procedure combining GMM and rank fallback is developed. This procedure identifies low-scoring neurons from the intra-layer neuron target score distribution, falls back to rank-based estimation when GMM evidence is insufficient, and maps weak-adaptation signals to selective soft decay factors. Unlike hard pruning, random resetting, or pure magnitude-based pruning, TGSR-PINN preserves network topology and applies selective soft decay only to the weight rows and biases corresponding to low-scoring neurons.
\item The proposed method is evaluated on high-P\'{e}clet number two-dimensional advection--diffusion, Allen--Cahn to Burgers cross-PDE-family transfer, and a 5\%-noise reaction--diffusion inverse problem. Mechanism ablation, scoring component scanning, GMM/rank fallback diagnosis, layer-protection analysis, overall intensity matching, selective soft decay mapping ablation, and computational overhead evaluation are further conducted to clarify the empirical advantages and applicability boundaries of TGSR-PINN.
\end{enumerate}

The remainder of this paper is organized as follows. Section~\ref{sec:method} presents the PDE inverse problem formulation and the overall framework of TGSR-PINN. Section~\ref{sec:experiments} presents the experimental design and results, including high-P\'{e}clet advection--diffusion, cross-PDE-family transfer, 5\%-noise reaction--diffusion experiments, and \revision{mechanism ablations on the smooth manufactured-solution advection--diffusion task}. Section~\ref{sec:discussion} discusses the relationship between field and parameter errors, the selective soft decay mechanism, computational overhead, and applicability boundaries. Section~\ref{sec:conclusion} concludes the paper and outlines future research directions.

\section{Method}
\label{sec:method}

\begin{figure}[tbp]
  \centering
  \includegraphics[width=.90\columnwidth,trim=0 0 0 10bp,clip]{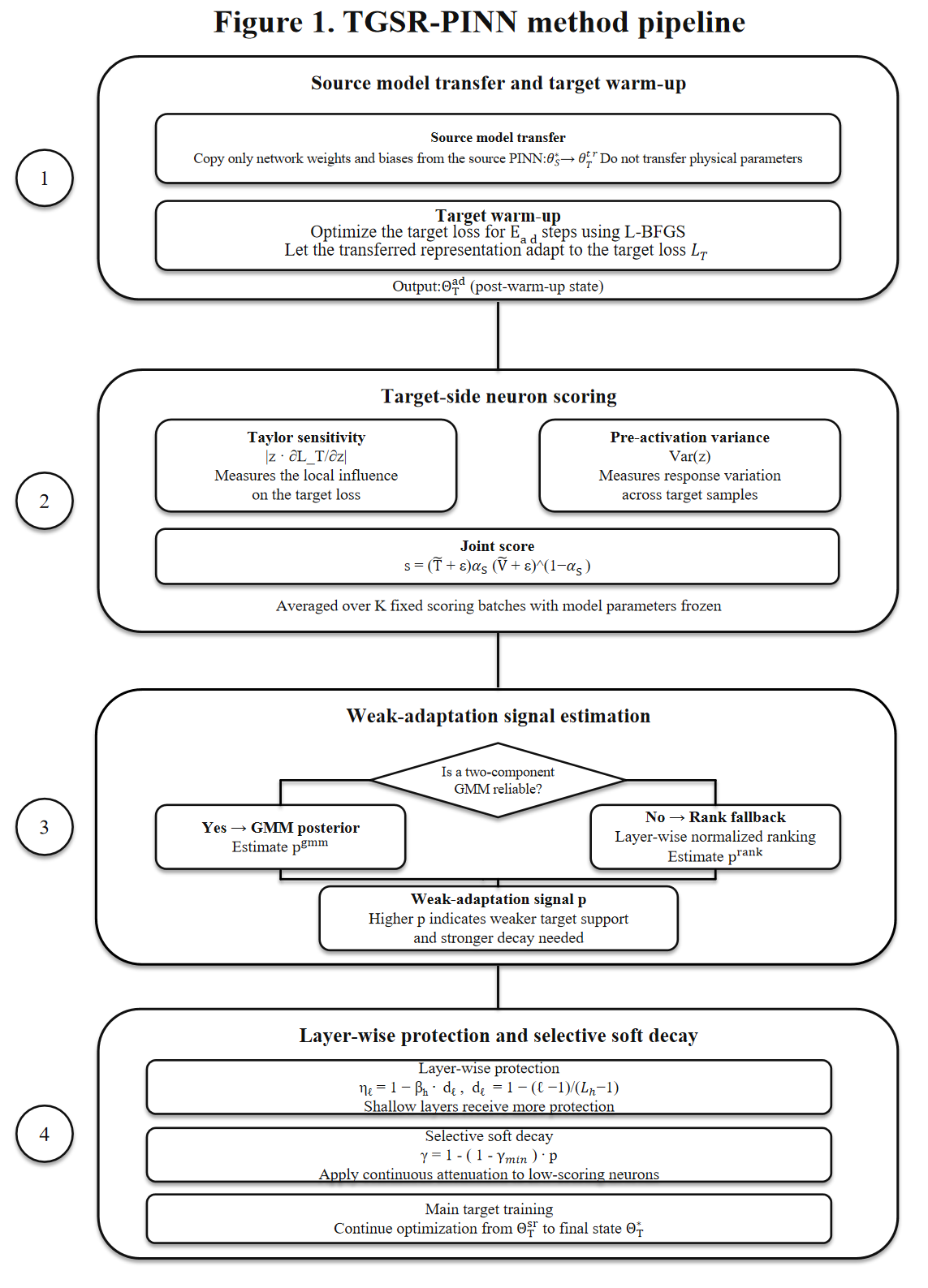}
  \caption{\revision{TGSR-PINN method pipeline.}}
  \label{fig:flowchart}
\end{figure}

\begin{revisionblock}
Figure~\ref{fig:flowchart} illustrates the TGSR-PINN pipeline.
\end{revisionblock}

\subsection{PINN Inverse Problem Formulation}
\label{subsec:pinn_inverse}

We first present the general PINN formulation for PDE inverse problems. Consider a physical system defined on a spatial domain $\Omega$ and a temporal interval $[0,\revision{T_f}]$, \revision{where $T_f$ is the final time}. We directly use spatial coordinates $x$ and time $t$ as network inputs; for steady-state problems, the time variable $t$ can be omitted. Let $u(x,t)$ denote the physical field to be solved and $\lambda$ denote the physical parameters to be inverted. The general governing equation, boundary conditions, and initial conditions can be written as:
\begin{equation}
  \mathcal{N}\!\left[ u(x,t);\lambda \right] = f(x,t),
  \label{eq:pde}
\end{equation}
\begin{equation}
  \mathcal{B}\!\left[ u(x,t);\lambda \right] = g(x,t),
  \label{eq:bc}
\end{equation}
\begin{equation}
  u(x,0) = u_{0}(x),
  \label{eq:ic}
\end{equation}
where Eq.~(\ref{eq:pde}) holds on $\Omega \times (0,\revision{T_f}]$, Eq.~(\ref{eq:bc}) holds on $\partial\Omega \times [0,\revision{T_f}]$, and Eq.~(\ref{eq:ic}) gives the initial state. $\mathcal{N}[\cdot]$ denotes the PDE differential operator, $\mathcal{B}[\cdot]$ denotes the boundary operator, $f(x,t)$ is the source term, and $g(x,t)$ and $u_{0}(x)$ denote boundary conditions and initial conditions, respectively. In inverse problems, $\lambda$ is the unknown parameter to be identified from limited observational data and physical constraints, which may represent diffusion coefficients, reaction coefficients, velocity parameters, material properties, or other physical quantities.

PINNs approximate the unknown physical field $u(x,t)$ using a fully connected neural network $u_{\theta}(x,t)$, where $\theta$ denotes the network parameters including weight matrices and bias vectors for each layer. For the target inverse problem, the physical parameters to be inverted are denoted $\lambda_{T}$, and the target network parameters are denoted $\theta_{T}$. Together they constitute the trainable variables $\Theta_{T}=\{\theta_{T},\lambda_{T}\}$ in the target task and are simultaneously updated under the target loss function.

Substituting the neural network prediction $u_{\theta_{T}}(x,t)$ into the governing equation yields the PDE residual for the target task:
\begin{equation}
  r\!\left( x,t;\Theta_{T} \right) = \mathcal{N}\!\left[ u_{\theta_{T}}(x,t);\lambda_{T} \right] - f(x,t).
  \label{eq:residual}
\end{equation}

The total loss for the target inverse problem is a weighted sum of components:
\begin{equation}
  \begin{aligned}
  \mathcal{L}_{T}\!\left( \Theta_{T} \right)
  &= \omega_{pde}\mathcal{L}_{pde}
   + \omega_{ic}\mathcal{L}_{ic} \\
  &\quad + \omega_{bc}\mathcal{L}_{bc}
   + \omega_{data}\mathcal{L}_{data}.
  \end{aligned}
  \label{eq:loss}
\end{equation}
where $\omega_{pde}$, $\omega_{ic}$, $\omega_{bc}$, and $\omega_{data}$ are the weights for each loss term. The focus of this paper is on how the transferred network representation is continuously corrected based on target-task evidence. Section~\ref{subsec:tgsr} will describe how TGSR-PINN reduces the influence of low-scoring neurons on target optimization through selective soft decay, thereby avoiding the excessive perturbation caused by random resetting.

\begin{revisionblock}
Parameter compensation is a recognized source of non-identifiability in inverse problems~\cite{daly2018identifiability}. In the joint PINN parameterization, let $F(\theta,\lambda)$ collect the weighted PDE, boundary/initial, and observation residuals near the current state. A first-order perturbation gives
\begin{equation}
  \Delta F \approx J_{\theta}\Delta\theta+J_{\lambda}\Delta\lambda,
  \qquad
  J_{\theta}=\frac{\partial F}{\partial\theta},\quad
  J_{\lambda}=\frac{\partial F}{\partial\lambda}.
  \label{eq:compensation_linearization}
\end{equation}
If $J_{\lambda}\Delta\lambda$ lies largely in the column space of $J_{\theta}$, the local direction
\begin{equation}
  \Delta\theta\approx-J_{\theta}^{\dagger}J_{\lambda}\Delta\lambda
  \label{eq:compensation_direction}
\end{equation}
can offset part of the residual change caused by an incorrect physical parameter, where $J_{\theta}^{\dagger}$ denotes the Moore--Penrose pseudoinverse. This relation is consistent with Jacobian-based analyses of PINN conditioning~\cite{cao2025illconditioning,rathore2024} and explains why a small field or residual error need not imply accurate parameter recovery. For the scalar residual $F_{\mathrm{ex}}(\theta,\lambda)=\theta+\lambda-y$, \revision{where $y$ is an observed scalar response}, the direction $\Delta\theta=-\Delta\lambda$ leaves the residual unchanged. This example illustrates local compensation; structural identifiability remains determined by the governing model and target observations.
\end{revisionblock}

\subsection{Overall Framework of TGSR-PINN}
\label{subsec:tgsr}

\begin{revisionblock}
TGSR-PINN stands for \textbf{Target-Guided Selective Reweighting PINN}. Target-Guided means that the scoring evidence is obtained from target losses, batches, gradients, and pre-activation responses, while Selective Reweighting means that the scaling strength varies across neurons. The neuron target score combines Taylor sensitivity and pre-activation variance, and the resulting weak-adaptation signal determines a bounded correction strength. All corrected neurons remain trainable.
\end{revisionblock}

\subsubsection*{Notation Summary}

\begin{table*}[!t]
\centering
\caption{Summary of main notation}
\label{tab:symbols}
\footnotesize
\setlength{\tabcolsep}{4pt}
\begin{tabularx}{\textwidth}{@{}lXlX@{}}
\toprule
Symbol & Meaning & Symbol & Meaning \\
\midrule
$\theta_{S}^{*}$ & Source representation parameters after source training
  & $s_{\ell,j}^{(k)},\;s_{\ell,j}$ & Per-batch and averaged neuron target score \\
\revision{$\theta,\;\lambda$} & \revision{Network parameters and unknown physical parameters}
  & \revision{$\widetilde{T}_{\ell,j}^{(k)},\;\widetilde{V}_{\ell,j}^{(k)}$} & \revision{Normalized Taylor and variance evidence} \\
\revision{$\Omega,\;T_f$} & \revision{Spatial domain and final time}
  & \revision{$u_{\theta_T},\;\Theta_T$} & \revision{Target field surrogate and joint trainable state} \\
\revision{$F,\;J_{\theta},\;J_{\lambda}$} & \revision{Joint residual vector and its two Jacobian blocks}
  & \revision{$\alpha_{\ell},\;L_h$} & \revision{Layer fusion weight and number of hidden layers} \\
\revision{$r(x,t;\Theta_T)$} & \revision{Target PDE residual}
  & \revision{$\psi_{\ell},\;\mathcal{G}$} & \revision{Layer-wise mixture density and Gaussian density} \\
$\theta_{T}^{tr}$ & Target network parameters copied from the source
  & \revision{$\chi_{\ell,j}$} & \revision{Compressed intra-layer neuron target score} \\
$\lambda_{T}^{0}$ & Independently initialized target physical parameters
  & $q_{\ell,j}^{L}$ & Posterior responsibility to the GMM low-mean component \\
$\Theta_{T}^{0}$ & Initial target state after representation transfer
  & $p_{\ell,j},\;\widetilde{p}_{\ell,j}$ & Weak-adaptation signal before/after layer protection \\
$\Theta_{T}^{ad}$ & Target state after short adaptation
  & $c_{\ell}$ & GMM two-component separation confidence \\
$\Theta_{T}^{sr}$ & Target state after selective soft decay
  & $\beta_h,\;\eta_{\ell},\;d_{\ell}$ & Protection strength, coefficient, and layer-position factor \\
$\Theta_{T}^{*}$ & Final target state after main training
  & $\mathcal{M}_{\ell}$ & Diagnostic set of effectively softened neurons \\
$\mathcal{L}_{T}$ & Target inverse loss
  & $\gamma_{\ell,j},\;\gamma_m,\;\gamma_{\min}$ & Neuron, intermediate, and minimum scaling factors \\
$E_{ad},\;E_{main}$ & Short-adaptation and main-training steps
  & $W_{\ell,j:},\;b_{\ell,j}$ & Input weight row and bias of neuron $(\ell,j)$ \\
$K,\;B_k^{score}$ & Number of scoring batches and the $k$-th batch
  & $\alpha_s,\;\Delta_\alpha$ & Base fusion weight and layer-wise increment \\
\revision{$\mathcal{B}_{score},\;\upsilon_{\ell,j}$} & \revision{Fixed scoring-batch set and virtual neuron gate}
  & \revision{$\tau_{\ell}$} & \revision{Robust intra-layer score scale} \\
$z_{\ell,j}^{(k)}$ & Neuron pre-activation on scoring batch $k$
  & \revision{$\pi_{\ell,a},\;\mu_{\ell,a},\;\sigma_{\ell,a}$} & \revision{GMM weight, mean, and standard deviation, $a\in\{L,H\}$} \\
$T_{\ell,j}^{(k)},\;V_{\ell,j}^{(k)}$ & Taylor sensitivity and pre-activation variance
  & \revision{$p_{\ell,j}^{gmm},\;p_{\ell,j}^{rank}$} & \revision{GMM and rank weak-adaptation signals} \\
\revision{$n_{\ell},\;\mathrm{rank}_{\ell}$} & \revision{Layer width and ascending score-rank operator}
  & \revision{$\xi,\;\rho_r,\;\delta_g$} & \revision{Rank-signal maximum, rank exponent, and reporting threshold} \\
\revision{$\kappa$} & \revision{Selective soft-decay mapping exponent}
  & \revision{$\vartheta_{\ell,j},\;L_{\vartheta}$} & \revision{Neuron parameter block and local gradient Lipschitz constant} \\
\revision{$\ell,\;j,\;k$} & \revision{Hidden-layer, neuron, and scoring-batch indices}
  & \revision{$\Delta\mathrm{BIC},\;BC_{\ell}$} & \revision{GMM model-selection gap and component-overlap coefficient} \\
\bottomrule
\end{tabularx}
\end{table*}

\revision{The table lists recurring method symbols. Auxiliary quantities used in a single derivation and task-specific physical coefficients are defined at first use. Numerical stability constants are denoted by $\zeta_0=10^{-6}$ in Eq.~(\ref{eq:score_single}) and $\zeta_s=\zeta_p=\zeta_r=10^{-12}$ in Eqs.~(\ref{eq:compress}), (\ref{eq:gmm_signal}), and (\ref{eq:rank}), respectively. The GMM variance floor is $10^{-10}$.}

TGSR-PINN starts from the following transfer setting: the source model provides only neural network representation parameters, and target physical parameters are independently initialized by the target task. This setting follows the fundamental principle in transfer learning of reusing transferable representations while avoiding direct inheritance of mismatched task states. Source-task physical parameters, optimizer states, training history, and sampling batches are not transferred. Let $\theta_{S}^{*}$ denote the network parameters after source training; then the target network is initialized as
\begin{equation}
  \theta_{T}^{tr} \leftarrow \theta_{S}^{*},\quad
  \Theta_{T}^{0} = \{\theta_{T}^{tr},\lambda_{T}^{0}\}.
  \label{eq:init}
\end{equation}
where $\lambda_{T}^{0}$ is the initial value of the target physical parameter. This setting ensures that the comparison focuses on transferred representations and their correction mechanisms, rather than direct inheritance of source-task physical parameters.

Subsequently, starting from $\Theta_{T}^{0}$, the model performs $E_{ad}$ steps of target short adaptation under the target loss $\mathcal{L}_{T}$. This stage simultaneously updates the transferred network parameters $\theta_{T}^{tr}$ and the target physical parameters $\lambda_{T}^{0}$, yielding the post-adaptation state:
\begin{equation}
  \Theta_{T}^{ad} = \{\theta_{T}^{ad},\lambda_{T}^{ad}\} = \mathrm{Train}\!\left( \Theta_{T}^{0},\mathcal{L}_{T},E_{ad} \right).
  \label{eq:adapt}
\end{equation}
\revision{Here $\mathrm{Train}(\Theta,\mathcal{L},E)$ denotes optimization of state $\Theta$ under loss $\mathcal{L}$ for $E$ training steps.}
\begin{revisionblock}
Target short adaptation exposes the copied network to target observations, boundary and initial conditions, and PDE constraints before neuron scoring. The score is therefore computed at $\Theta_{T}^{ad}$, after the model has acquired preliminary target responses. It is not computed directly from the source state. The adaptation length is set by a proportion of the target training budget and an upper bound. An excessively long adaptation stage or severely misaligned initial parameters can contaminate the score through early parameter compensation. We therefore control the adaptation length and examine its sensitivity experimentally.
\end{revisionblock}

After completing target short adaptation, TGSR-PINN temporarily fixes $K$ scoring batches from the target training sampler:
\begin{equation}
  \mathcal{B}_{score} = \{ B_{1}^{score},B_{2}^{score},\ldots,B_{K}^{score}\}.
  \label{eq:batch}
\end{equation}
Each scoring batch is consistent with the sample types in the target loss, including PDE collocation points, boundary condition points, initial condition points, and observation data points. If a task does not include a certain constraint type, the scoring batch does not include the corresponding samples either. Unless overridden by configuration, the experimental implementation defaults to $K=3$. \revision{During scoring, the post-adaptation network parameters $\theta_{T}^{ad}$ and target physical parameters $\lambda_{T}^{ad}$ are frozen at $\Theta_{T}^{ad}$.} This phase performs only forward computation and gradient backpropagation to collect hidden neuron pre-activation responses, target loss gradients, and pre-activation variance information; no optimizer updates are executed. After scoring, these cached batches no longer participate as fixed constraints in subsequent training, and the model continues sampling and optimizing according to the original target training protocol.

Consider the $j$-th hidden neuron in the $\ell$-th layer. In our implementation, hooks are registered at the linear output of each fully connected layer, so we denote its pre-activation response on the $k$-th scoring batch as $z_{\ell,j}^{(k)}(x,t)$. Let $\mathcal{L}_{T}^{(k)}$ denote the target loss evaluated on that batch. To approximately measure the neuron's local contribution to this loss, we introduce a virtual gating variable \revision{$\upsilon_{\ell,j}$}, writing the pre-activation as \revision{$\upsilon_{\ell,j}z_{\ell,j}^{(k)}(x,t)$}. During normal forward computation, \revision{$\upsilon_{\ell,j}=1$}; perturbing it to 0 can be approximately understood as suppressing that neuron's input response. This scoring idea is consistent with Taylor-expansion-based neuron importance estimation for pruning~\cite{molchanov2019}. Based on the first-order Taylor expansion, the Taylor sensitivity on the $k$-th scoring batch is defined as
\begin{equation}
  T_{\ell,j}^{(k)} = \mathbb{E}_{(x,t) \in B_{k}^{score}}\left| z_{\ell,j}^{(k)}(x,t)\frac{\partial\mathcal{L}_{T}^{(k)}}{\partial z_{\ell,j}^{(k)}(x,t)} \right|.
  \label{eq:taylor}
\end{equation}

\begin{revisionblock}
More formally, let \revision{$\phi_{\ell,j}^{(k)}(\upsilon)$} denote the target loss when only this pre-activation is replaced by \revision{$\upsilon z_{\ell,j}^{(k)}$}. For a small attenuation $\delta_a\in[0,1]$, Taylor's theorem gives
\begin{equation}
  \phi_{\ell,j}^{(k)}(1-\delta_a)-\phi_{\ell,j}^{(k)}(1)
  =-\delta_a\,\phi_{\ell,j}^{(k)\prime}(1)+R_{\ell,j}^{(k)}(\delta_a),
  \label{eq:gate_taylor}
\end{equation}
where $R_{\ell,j}^{(k)}$ is the Taylor remainder, $\phi_{\ell,j}^{(k)\prime}(1)=\mathbb{E}[z_{\ell,j}^{(k)}\partial\mathcal{L}_{T}^{(k)}/\partial z_{\ell,j}^{(k)}]$, and $|\phi_{\ell,j}^{(k)\prime}(1)|\leq T_{\ell,j}^{(k)}$ by the triangle inequality. If \revision{$|\phi_{\ell,j}^{(k)\prime\prime}(\upsilon)|\leq M_{\ell,j}^{(k)}$ on $\upsilon\in[1-\delta_a,1]$}, where $M_{\ell,j}^{(k)}$ is a local curvature bound, then
\begin{equation}
  |R_{\ell,j}^{(k)}(\delta_a)|\leq \tfrac{1}{2}M_{\ell,j}^{(k)}\delta_a^2.
  \label{eq:gate_remainder}
\end{equation}
Thus, under a local gate perturbation, Eq.~(\ref{eq:taylor}) approximates the target-loss change caused by neuron attenuation, consistent with Taylor importance estimation~\cite{molchanov2019,sanh2020}. The remainder bound in Eq.~(\ref{eq:gate_remainder}) describes the approximation error, while the explicit gating diagnostic in Section~\ref{subsec:ablation} provides experimental support by comparing the score with the actual loss response.
\end{revisionblock}

Taylor sensitivity reflects the first-order sensitivity of the target loss to this neuron's pre-activation response. A larger value indicates that changes in the neuron's pre-activation have a more pronounced effect on the current target loss; the absolute value in Eq.~(\ref{eq:taylor}) prevents positive and negative gradient directions from canceling each other.

Relying solely on Taylor sensitivity may be affected by local gradient fluctuations. Some neurons may have large gradients on specific batches but weak response variation on target samples. Therefore, TGSR-PINN simultaneously computes pre-activation variance:
\begin{equation}
  V_{\ell,j}^{(k)} = \mathrm{Var}_{(x,t) \in B_{k}^{score}}\!\left( z_{\ell,j}^{(k)}(x,t) \right).
  \label{eq:variance}
\end{equation}
\begin{revisionblock}
Under the same gate perturbation, $\Delta z_{\ell,j}^{(k)}=-\delta_a z_{\ell,j}^{(k)}$ and therefore
\begin{equation}
  \mathrm{Var}(\Delta z_{\ell,j}^{(k)})=\delta_a^2 V_{\ell,j}^{(k)}.
  \label{eq:variance_gate}
\end{equation}
Consequently, $V_{\ell,j}^{(k)}$ describes how the neuron response varies across target samples, whereas $T_{\ell,j}^{(k)}$ describes the local sensitivity of the target loss to that response. The two quantities characterize different aspects of the adapted neuron state. Section~\ref{subsec:ablation} examines how their relative weighting affects parameter recovery in the tested task.
\end{revisionblock}
Here, the variance specifically refers to the pre-activation variance of the linear layer output $z_{\ell,j}$, not the variance of the post-activation value $\tanh(z_{\ell,j})$. We use pre-activation statistics because the hooks in our implementation are registered at the linear layer output, and subsequent selective soft decay directly modulates the pre-activation magnitude by scaling $W_{\ell,j:}$ and $b_{\ell,j}$. Deep network initialization and normalization research has long focused on the mean and variance of layer inputs or neuron summed inputs to maintain signal propagation and training stability~\cite{glorot2010,ioffe2015,ba2016}. Thus, pre-activation variance characterizes the linear response variation range of the neuron on target samples and serves as an auxiliary statistic for target sample response activity in neuron target scoring.

Since numerical scales may differ across layers, we first normalize $T_{\ell,j}^{(k)}$ and $V_{\ell,j}^{(k)}$ within each layer separately, obtaining $\widetilde{T}_{\ell,j}^{(k)}$ and $\widetilde{V}_{\ell,j}^{(k)}$. The implementation uses layer-wise z-score, negative-value truncation, and maximum rescaling to bring both types of evidence into $[0,1]$. Let $L_h$ denote the number of hidden layers. Before geometric fusion, we define $\alpha_{\ell}=\alpha_s+\Delta_{\alpha}(\ell-1)/(L_h-1)$ to control the relative weight of Taylor sensitivity and pre-activation variance in layer $\ell$; for a single hidden layer, we set $\alpha_{\ell}=\alpha_s+\Delta_{\alpha}/2$. We then compute the neuron target score for this batch using geometric fusion:
\begin{equation}
  s_{\ell,j}^{(k)} = \left( \widetilde{T}_{\ell,j}^{(k)} + \revision{\zeta_0} \right)^{\alpha_{\ell}}\!\left( \widetilde{V}_{\ell,j}^{(k)} + \revision{\zeta_0} \right)^{1 - \alpha_{\ell}}.
  \label{eq:score_single}
\end{equation}
where $\zeta_0=10^{-6}$ is the numerical stability term in Eq.~(\ref{eq:score_single}). The experimental defaults are $\alpha_s=0.5$ and $\Delta_{\alpha}=0.3$, so \revision{deeper-layer scores place greater weight on Taylor sensitivity}. Compared with arithmetic averaging, this geometric fusion is more sensitive to the smaller of the two evidence terms: if either Taylor sensitivity or pre-activation variance is weak, the neuron target score is suppressed. \revision{This reduces the risk of assigning strong target relevance from only a large gradient or response magnitude.} Averaging over $K$ scoring batches yields the neuron target score for neuron $j$ in layer $\ell$:
\begin{revisionblock}
The geometric fusion is monotone and conservative. A low value in either normalized evidence channel reduces the fused score. The resulting score serves as a local target-relevance proxy at the evaluated adaptation state. Section~\ref{subsec:ablation} examines the fusion weight and its consistency with local perturbations.
\end{revisionblock}
\begin{equation}
  s_{\ell,j} = \frac{1}{K}\sum_{k = 1}^{K}s_{\ell,j}^{(k)}.
  \label{eq:score_avg}
\end{equation}
A higher $s_{\ell,j}$ indicates stronger target evidence supporting retention of the neuron; a lower neuron target score indicates weaker target evidence support, and the subsequent continuous weak-adaptation signal will determine the selective soft decay intensity.

After obtaining neuron target scores $s_{\ell,j}$, TGSR-PINN avoids both a fixed numerical score threshold and a fixed neuron-intervention ratio because score scales and distribution shapes vary significantly across layers. We convert neuron target scores to weak-adaptation signals $p_{\ell,j}$: higher $s_{\ell,j}$ means the current target evidence better supports retaining the neuron, while higher $p_{\ell,j}$ means the neuron requires more weakening. To reduce the influence of extreme high scores on distribution estimation, we first apply a compression transform to each layer's neuron target scores:
\begin{equation}
  \begin{aligned}
  \revision{\tau_{\ell}} &=
  \mathrm{median}\!\left\{s_{\ell,j}:s_{\ell,j}>0\right\}
  + \revision{\zeta_{s}},\\
  \revision{\chi_{\ell,j}} &=
  \log\!\left(1+\frac{s_{\ell,j}}{\revision{\tau_{\ell}}}\right).
  \end{aligned}
  \label{eq:compress}
\end{equation}
where $\tau_{\ell}$ is the robust score scale for layer $\ell$ and $\zeta_{s}=10^{-12}$ is a numerical stability term.

Subsequently, a one-dimensional two-component GMM is fit on each layer's \revision{$\chi_{\ell,j}$}, with parameters estimated via the EM algorithm~\cite{dempster1977}:
\begin{equation}
  \begin{aligned}
  \revision{\psi_{\ell}(\chi)}
  &= \pi_{\ell,L}
     \revision{\mathcal{G}}\!\left( \chi;\mu_{\ell,L},\sigma_{\ell,L}^{2} \right)\\
  &\quad + \pi_{\ell,H}
     \revision{\mathcal{G}}\!\left( \chi;\mu_{\ell,H},\sigma_{\ell,H}^{2} \right).
  \end{aligned}
  \label{eq:gmm}
\end{equation}
where \revision{$\psi_{\ell}(\chi)$ is the layer-wise mixture density and $\mathcal{G}(\chi;\mu,\sigma^2)$ denotes a Gaussian density}; the lower-mean component $L$ represents the candidate low-scoring-neuron component, and the higher-mean component $H$ represents the relatively high-scoring-neuron component. For neuron $j$, its posterior responsibility to the low-mean component is
\begin{equation}
  q_{\ell,j}^{L} = \frac{\pi_{\ell,L}\revision{\mathcal{G}}\!\left( \revision{\chi_{\ell,j}};\mu_{\ell,L},\sigma_{\ell,L}^{2} \right)}{\revision{\psi_{\ell}}\!\left( \revision{\chi_{\ell,j}} \right)}.
  \label{eq:responsibility}
\end{equation}

To make it more suitable for subsequent selective soft decay, we calibrate it to obtain the GMM branch's weak-adaptation signal:
\begin{equation}
  p_{\ell,j}^{gmm} = c_{\ell}\cdot \mathrm{clip}\!\left( \frac{q_{\ell,j}^{L} - \pi_{\ell,L}}{1 - \pi_{\ell,L} + \revision{\zeta_{p}}},0,1 \right).
  \label{eq:gmm_signal}
\end{equation}
where $\mathrm{clip}(v,0,1)=\min\{1,\max\{0,v\}\}$, $c_{\ell}\in[0,1]$ is the separation confidence derived from the two-component overlap, and $\zeta_p=10^{-12}$ is a numerical stability term. In implementation, $c_{\ell}=1-BC_{\ell}$, where $BC_{\ell}$ is the Bhattacharyya coefficient~\cite{bhattacharyya1943} between the two Gaussian components. The stronger the overlap between the two components, the smaller $c_{\ell}$ becomes, reducing the weak-adaptation signal from an uncertain GMM partition. In Eq.~(\ref{eq:gmm_signal}), $q_{\ell,j}^{L}-\pi_{\ell,L}$ denotes the posterior excess of the low-mean component responsibility over its layer-wise prior weight. Only posterior evidence exceeding this prior is converted into a weak-adaptation signal, which avoids interpreting a large low-scoring component as an anomalous subgroup requiring strong selective soft decay. Thus, the GMM branch derives layer-specific grouping from distributional evidence when a \revision{non-majority} low-score component is supported, while rank fallback maintains a bounded monotone signal when that evidence is insufficient.

For implementation, TGSR-PINN fits the GMM to the compressed scores $\log(1+s/\tau_{\ell})$ in Eq.~(\ref{eq:compress}). Here $\pi_{\ell,a}$, $\mu_{\ell,a}$, and $\sigma_{\ell,a}$ denote the mixture weight, mean, and standard deviation of component $a\in\{L,H\}$. The component means are initialized at the 25th and 75th percentiles of the compressed scores, the mixture weights are initialized to 0.5, the variance floor is set to $10^{-10}$, and EM runs for at most 100 iterations with convergence declared when the log-likelihood change is below $10^{-8}$. We define $\Delta \mathrm{BIC}=\mathrm{BIC}_{1G}-\mathrm{BIC}_{2G}$, where $\mathrm{BIC}_{1G}$ is the BIC of the single-Gaussian model and $\mathrm{BIC}_{2G}$ is the BIC of the two-component GMM. Since lower BIC is better, $\Delta \mathrm{BIC}>0$ indicates that the two-component model is preferred over the single-Gaussian model. In implementation, we use $\Delta \mathrm{BIC}>10$ and $\pi_{\ell,L}\leq 0.5$ as the primary conditions for enabling the GMM branch~\cite{schwarz1978}. The constraint $\pi_{\ell,L}\leq 0.5$ ensures that the low-mean component \revision{does not constitute a majority of the layer-wise mixture mass}. If the conditions are not met, the GMM branch is abandoned and rank fallback is activated:
\begin{equation}
  p_{\ell,j}^{rank} = \xi \cdot \left(\frac{n_{\ell} - \mathrm{rank}_{\ell}(s_{\ell,j})}{n_{\ell} - 1 + \revision{\zeta_{r}}}\right)^{\rho_{r}}.
  \label{eq:rank}
\end{equation}
where $n_{\ell}$ is the width of layer $\ell$, $\mathrm{rank}_{\ell}(s_{\ell,j})$ is the ascending rank of $s_{\ell,j}$ within that layer, $\xi=0.75$, $\rho_r=1.5$, and $\zeta_r=10^{-12}$.

To prevent excessive intervention on shallow input representations, we first apply layer protection to the weak-adaptation signals. With $L_{h}$ hidden layers:
\begin{equation}
  \begin{aligned}
  d_{\ell}
  &=
  1-\frac{\ell-1}{L_{h}-1},\\
  \eta_{\ell}
  &=
  1-\beta_{h}d_{\ell},\\
  \widetilde{p}_{\ell,j}
  &=
  \mathrm{clip}\!\left(\eta_{\ell}p_{\ell,j},0,1\right).
  \end{aligned}
  \label{eq:layer_protect}
\end{equation}
where $\beta_h=0.55$ is the shallow protection strength~\cite{yosinski2014}. For a single-hidden-layer network, we set $d_{\ell}=1$.

Based on the adjusted weak-adaptation signal $\widetilde{p}_{\ell,j}$, TGSR-PINN maps it directly to a continuous scaling coefficient:
\begin{equation}
  \begin{aligned}
  \gamma_{\ell,j}
  &=
  \begin{cases}
    1 - A_{\ell,j}(1-\gamma_{m}),
    & \widetilde{p}_{\ell,j}<0.5,\\
    \gamma_{m} - B_{\ell,j}(\gamma_{m}-\gamma_{\min}),
    & \widetilde{p}_{\ell,j}\geq 0.5,
  \end{cases}\\
  A_{\ell,j}
  &= \left(2\widetilde{p}_{\ell,j}\right)^{\revision{\kappa}},\\
  B_{\ell,j}
  &= \left(2\widetilde{p}_{\ell,j}-1\right)^{\revision{\kappa}}.
  \end{aligned}
  \label{eq:decay_map}
\end{equation}
where $0\leq\widetilde{p}_{\ell,j}\leq1$, $\gamma_{m}=0.85$, $\gamma_{\min}=0.4$, and the default mapping exponent is $\kappa=3$. This mapping is used as a monotone, continuous, and lower-bounded selective scaling rule; it provides mild selective soft decay for moderate weak-adaptation signals and stronger but still recoverable selective soft decay for pronounced weak-adaptation signals. Its motivation is consistent with soft pruning and soft-threshold sparsification, which preserve network capacity and avoid irreversible deletion~\cite{he2018soft,kusupati2020,sanh2020}. Section~\ref{subsec:ablation} further compares this default mapping with linear, sigmoid, and hard-threshold alternatives.

\begin{revisionblock}
The GMM and rank branches derive $p_{\ell,j}$ differently, but both map it to $[0,1]$ with the same monotone interpretation. A larger value requests stronger attenuation. Equation~(\ref{eq:decay_map}) satisfies $\gamma(0)=1$, $\gamma(0.5)=\gamma_m$, and $\gamma(1)=\gamma_{\min}$. The value 0.5 is a knot joining mild- and stronger-attenuation regimes. It does not select, remove, or freeze neurons. Every neuron receives a continuous scale and remains trainable. With the default $\kappa=3$, attenuation increases slowly for low and moderate signals and more rapidly for pronounced signals. This shape reduces sensitivity to small score fluctuations. Differentiability at the knot is unnecessary because the mapping is applied once between adaptation and main training; it is not optimized as a persistent network module.
\end{revisionblock}

For reporting and visualization, we define the diagnostic set of effectively softened neurons as
\begin{equation}
  \mathcal{M}_{\ell}=\{j:1-\gamma_{\ell,j}\geq\delta_g\},
  \label{eq:softened_set}
\end{equation}
where $\delta_g=0.02$ is only a reporting threshold. The actual operation is still the continuous scaling by $\gamma_{\ell,j}$, rather than hard deletion or random resetting based on $\mathcal{M}_{\ell}$.

Selective soft decay is written as
\begin{equation}
  W_{\ell,j:} \leftarrow \gamma_{\ell,j}W_{\ell,j:},\quad
  b_{\ell,j} \leftarrow \gamma_{\ell,j}b_{\ell,j}.
  \label{eq:soft_decay}
\end{equation}
Let $z_{\ell,j}=W_{\ell,j:}h_{\ell-1}+b_{\ell,j}$. At the instant when Eq.~(\ref{eq:soft_decay}) is applied, for a fixed previous-layer input $h_{\ell-1}$, the updated pre-activation satisfies $z'_{\ell,j}=\gamma_{\ell,j}z_{\ell,j}$. Therefore, Eq.~(\ref{eq:soft_decay}) is consistent with the virtual pre-activation gate used for scoring in Eq.~(\ref{eq:taylor}) at the pre-activation level: both correspond to multiplicative modulation of $z_{\ell,j}$. It should be distinguished from a post-activation gate $\gamma_{\ell,j}\tanh(z_{\ell,j})$ and from outgoing weight scaling in the next layer. Under nonlinear activations such as tanh, scaling the pre-activation is not exactly equivalent to linearly scaling the activation output. We adopt weight-row and bias scaling because it requires no additional network modules or permanent masks and matches the parameterization used in the implementation.

\begin{revisionblock}
Let $\vartheta_{\ell,j}=[W_{\ell,j:},b_{\ell,j}]$ denote the parameter block of neuron $(\ell,j)$ and $\Delta\vartheta_{\ell,j}=(\gamma_{\ell,j}-1)\vartheta_{\ell,j}$. The intervention obeys the deterministic bound
\begin{equation}
  \|\Delta\vartheta_{\ell,j}\|_2
  =(1-\gamma_{\ell,j})\|\vartheta_{\ell,j}\|_2
  \leq(1-\gamma_{\min})\|\vartheta_{\ell,j}\|_2.
  \label{eq:decay_bound}
\end{equation}
If the target loss has a locally $L_{\vartheta}$-Lipschitz gradient with respect to $\vartheta_{\ell,j}$, the descent lemma further gives
\begin{equation}
  \mathcal{L}_T(\vartheta+\Delta\vartheta)-\mathcal{L}_T(\vartheta)
  \leq \nabla_{\vartheta}\mathcal{L}_T(\vartheta)^{\top}\Delta\vartheta
  +\tfrac{L_{\vartheta}}{2}\|\Delta\vartheta\|_2^2.
  \label{eq:soft_loss_bound}
\end{equation}
Equations~(\ref{eq:decay_bound})--(\ref{eq:soft_loss_bound}) establish two properties of the intervention: the state change is continuous and bounded, and every scaled parameter remains trainable. The descent-lemma bound further characterizes the local loss variation induced by the scaling step. In contrast, a permanent hard mask sets $\gamma=0$ and removes the corresponding recovery direction, while random resetting lacks an analogous bound relative to the adapted state. This recoverability distinction is consistent with soft pruning and smooth thresholding methods that retain optimization freedom after attenuation~\cite{he2018soft,kusupati2020}.
\end{revisionblock}

Since $\gamma_{\min}>0$, Eq.~(\ref{eq:soft_decay}) does not remove neurons or alter network topology. The selectively decayed weights and biases remain trainable in subsequent main training, so low-scoring neurons can still recover if later target optimization makes them useful. Denoting the post-decay target state as
\begin{equation}
  \Theta_{T}^{sr}=\mathcal{S}_{sr}(\Theta_{T}^{ad};\{\gamma_{\ell,j}\}),
  \label{eq:state_sr}
\end{equation}
where $\mathcal{S}_{sr}$ denotes the selective-soft-decay state transformation. The model finally continues training to obtain:
\begin{equation}
  \Theta_{T}^{*} = \mathrm{Train}(\Theta_{T}^{sr},\mathcal{L}_{T},E_{main}).
  \label{eq:final}
\end{equation}
\subsection{Algorithm Overview}
\label{subsec:algorithm}

Algorithm~\ref{alg:tgsr} summarizes the main steps of TGSR-PINN.

\begin{algorithm}[h]
\footnotesize
\caption{TGSR-PINN Target-Guided Selective Reweighting Pipeline}
\label{alg:tgsr}
\KwIn{source network $\theta_S^*$, target parameter init $\lambda_T^0$, target loss $\mathcal{L}_T$, $E_{ad}$, $K$, $E_{main}$}
\KwOut{final target state $\Theta_T^*$}
Copy network weights only: $\theta_T^{tr}\leftarrow\theta_S^*$; initialize $\lambda_T^0$ independently (Eq.~\ref{eq:init})\;
Train $\{\theta_T^{tr},\lambda_T^0\}$ for $E_{ad}$ target short adaptation steps to obtain $\Theta_T^{ad}$ (Eq.~\ref{eq:adapt})\;
Fix $K$ target scoring batches $\mathcal{B}_{score}$ and freeze $\Theta_T^{ad}$ during scoring (Eq.~\ref{eq:batch})\;
\ForEach{hidden neuron $(\ell,j)$}{
  Compute $T_{\ell,j}^{(k)}$ and $V_{\ell,j}^{(k)}$ on each scoring batch (Eqs.~\ref{eq:taylor}--\ref{eq:variance})\;
  Normalize and fuse them to obtain $s_{\ell,j}^{(k)}$, then average over $K$ batches to obtain $s_{\ell,j}$ (Eqs.~\ref{eq:score_single}--\ref{eq:score_avg})\;
}
\ForEach{hidden layer $\ell$}{
  Compress intra-layer scores $s_{\ell,j}$ to \revision{$\chi_{\ell,j}$} and fit a two-component GMM (Eqs.~\ref{eq:compress}--\ref{eq:gmm})\;
  \eIf{$\Delta\mathrm{BIC}>10$ and $\pi_{\ell,L}\leq0.5$}{
    Compute GMM weak-adaptation signals $p_{\ell,j}^{gmm}$ (Eqs.~\ref{eq:responsibility}--\ref{eq:gmm_signal})\;
  }{
    Compute rank-fallback weak-adaptation signals $p_{\ell,j}^{rank}$ (Eq.~\ref{eq:rank})\;
  }
  Set $p_{\ell,j}$ from the active branch and apply layer protection to obtain $\widetilde{p}_{\ell,j}$ (Eq.~\ref{eq:layer_protect})\;
  Map $\widetilde{p}_{\ell,j}$ to $\gamma_{\ell,j}$ and scale $W_{\ell,j:}$ and $b_{\ell,j}$ (Eqs.~\ref{eq:decay_map}--\ref{eq:soft_decay})\;
}
Continue target main training from $\Theta_T^{sr}$ for $E_{main}$ steps to obtain $\Theta_T^*$ (Eqs.~\ref{eq:state_sr}--\ref{eq:final})\;
\Return $\Theta_T^*$\;
\end{algorithm}

\section{Experimental Design and Results}
\label{sec:experiments}

\subsection{Overall Protocol and Evaluation Metrics}
\label{subsec:protocol}

The baseline methods compared in this paper include:
\begin{itemize}
  \item \textbf{PINNs}: No source model; target task trained from random initialization;
  \item \revision{\textbf{Full} (full fine-tuning): All source weights and biases are copied and updated;}
  \item \textbf{Lightweight Fine-Tuning}: Earlier layers frozen, only later layers updated;
  \item \textbf{Partial Transfer}: Only selected network layers transferred;
  \item \textbf{TL-gPINN}: Transfer learning combined with gradient-enhanced PINN;
  \item \textbf{BitFit}: Only bias terms are trained.
  \item \revision{\textbf{L2-SP}: Target fine-tuning with an $L_2$ penalty anchoring transferred weights to the source representation;}
  \item \revision{\textbf{SFP}: Soft Filter Pruning~\cite{he2018soft}, applied to the hidden neurons of the PINN; magnitude-selected neurons are zeroed once but remain trainable during subsequent target optimization.}
\end{itemize}

Evaluation metrics include relative $L_2$ field error and average parameter error, both reported as percentages. The former represents the relative $L_2$ error of the predicted physical field with respect to the true solution, while the latter denotes the mean of the relative errors of the physical parameters to be inverted. For multi-parameter inversion tasks, the average parameter error is the arithmetic mean of the absolute relative errors of each physical parameter and is therefore non-negative. In the result tables, $n$ denotes the number of random-seed repetitions. Since field error and parameter error may not be consistent in PINN inverse problems, we report both metrics simultaneously and focus on parameter recovery quality in the main experiments.

\revision{All main experimental tasks use ten paired repetitions with different random seeds, and the main text reports aggregated results together with representative diagnostics. Within each repetition, all methods share the same seed-controlled initialization protocol, observations, collocation points, and training budgets.} During training, the LBFGS stage uses a fixed-sampling-batch early stopping rule: if the relative improvement in total loss falls below $10^{-8}$ for consecutive rounds, that repetition terminates early. This early stopping condition depends only on training loss and does not use true physical parameter errors or test field errors as stopping criteria, thus avoiding leakage of true label information into the training process.

\begin{revisionblock}
We use matched repetitions to control differences in random initialization, observation sampling, and optimization paths. Their final field and parameter errors are therefore treated as paired samples. The main results report the mean, standard deviation, paired win count, and Wilcoxon $p$ value across random-seed repetitions. Representative figures use one matched repetition for visualization only; all conclusions rely on the complete paired cohort.
\end{revisionblock}

\revision{For inferential comparisons under this matched design, we use two-sided paired Wilcoxon signed-rank tests and report the unadjusted $p$ values together with paired win counts. Differences with $p<0.05$ are treated as statistically significant within the corresponding comparison.}

\begin{revisionblock}
Negative transfer generally refers to degradation in target performance caused by transferred source knowledge~\cite{rosenstein2005,pan2010,wang2019negative}. Following the target-risk-gap formulation of Wang et al.~\cite{wang2019negative}, we quantify transfer effects relative to a matched target-only PINN baseline. For a transfer method $m$ and a matched repetition $i$, the target-side transfer effects are defined as
\begin{equation}
  \Delta_u^{(m,i)}=e_u^{(m,i)}-e_u^{(\mathrm{PINN},i)},\qquad
  \Delta_\lambda^{(m,i)}=e_\lambda^{(m,i)}-e_\lambda^{(\mathrm{PINN},i)},
  \label{eq:negative_transfer}
\end{equation}
where $e_u$ is the relative field error, $e_\lambda$ is the average physical-parameter error, $m$ indexes the transfer method, and $i$ indexes the matched repetition. Because parameter identification is the primary objective, parameter negative transfer is recorded when $\Delta_\lambda^{(m,i)}>0$. We further define a masked negative-transfer event as
\begin{equation}
  \Delta_u^{(m,i)}\leq 0 \quad\text{and}\quad
  \Delta_\lambda^{(m,i)}>0,
  \label{eq:masked_negative_transfer}
\end{equation}
This condition identifies cases in which transfer preserves or improves the field metric while degrading parameter recovery. Using matched standard PINN runs as the common reference, the event analysis compares \revision{Full} and TGSR-PINN, while the remaining methods are retained in the task-level performance comparisons.

\begin{table}[t]
\centering
\caption{\revision{Observed negative-transfer event counts.}}
\label{tab:negative_transfer_frequency}
\small
\begin{tabularx}{\columnwidth}{@{}lYYYY@{}}
\toprule
& \multicolumn{2}{c}{Parameter negative transfer} & \multicolumn{2}{c}{Masked events} \\
\cmidrule(lr){2-3}\cmidrule(lr){4-5}
Task & \revision{Full} & TGSR & \revision{Full} & TGSR \\
\midrule
Boundary layer & 7/10 & \textbf{0/10} & 0/10 & 0/10 \\
Cross-PDE & 6/10 & \textbf{3/10} & 4/10 & \textbf{2/10} \\
Reaction--diffusion & 8/10 & \textbf{1/10} & 3/10 & \textbf{1/10} \\
\bottomrule
\end{tabularx}
\end{table}
Across the three cohorts, the observed parameter-negative-transfer counts decrease from 7/10 to 0/10 on the boundary-layer task, from 6/10 to 3/10 on Cross-PDE transfer, and from 8/10 to 1/10 on reaction--diffusion inversion. Masked-event counts also decrease on the latter two tasks. These counts characterize the transfer behavior observed under the paired experimental protocol.
\end{revisionblock}

\begin{revisionblock}
\subsection{High-P\'{e}clet 2D Inflow--Outflow Boundary-Layer Inverse Problem}
\label{subsec:high_pe}

We consider a steady two-dimensional inflow--outflow problem with a zero source term, non-zero Dirichlet data, and an outflow boundary layer near $x=1$:
\begin{equation}
  -\epsilon\left(u_{xx}+u_{yy}\right)+\revision{a_x} u_x=0,
  \qquad (x,y)\in(0,1)^2.
  \label{eq:boundary_layer_pde}
\end{equation}
Here \revision{$a_x=1$ is the advection coefficient} and the unknown diffusion coefficient is $\epsilon$. With the characteristic length $L_c=1$, we use the convention \revision{$Pe=a_xL_c/\epsilon$}. The target value $\epsilon=0.01$ therefore gives $Pe=100$ and an outflow-layer width of order \revision{$\delta_{\mathrm{BL}}=\epsilon/a_x=0.01$}.

The analytical reference solution is
\begin{equation}
  u(x,y)=y\frac{1-\exp[-Pe(1-x)]}{1-\exp(-Pe)},
  \label{eq:boundary_layer_exact}
\end{equation}
which satisfies Eq.~\eqref{eq:boundary_layer_pde} without a forcing term. Its Dirichlet data are $u(0,y)=y$, $u(1,y)=0$, $u(x,0)=0$, and $u(x,1)=[1-\exp(-Pe(1-x))]/[1-\exp(-Pe)]$. Thus the solution drops sharply at the outflow boundary $x=1$ rather than remaining globally smooth.

\begin{table}[t]
\centering
\caption{High-P\'{e}clet boundary-layer inverse-problem settings.}
\label{tab:pe_setting}
\small
\begin{tabularx}{\columnwidth}{@{}lX@{}}
\toprule
Item & Setting \\
\midrule
Source task & Same PDE family, $Pe=5$ \\
Target task & Zero-source boundary layer, $Pe=100$ \\
Unknown / truth / init & $\epsilon$ / $0.01$ / $0.03$ \\
Boundary data & Non-zero inflow and top Dirichlet data \\
Noise & 0.5\% on target observations \\
Sampling & PDE 10000, BC 2500, data 1000 \\
Loss weights & $\omega_{pde}=10$, $\omega_{bc}=10$, $\omega_{data}=50$ \\
Network & MLP $[2,100^6,1]$, tanh \\
Optimizer & LBFGS only, at most 300 outer epochs \\
Local evaluation & Boundary-layer region $x\geq0.95$ \\
\bottomrule
\end{tabularx}
\end{table}

\begin{figure}[tbp]
  \centering
  \includegraphics[width=.94\columnwidth]{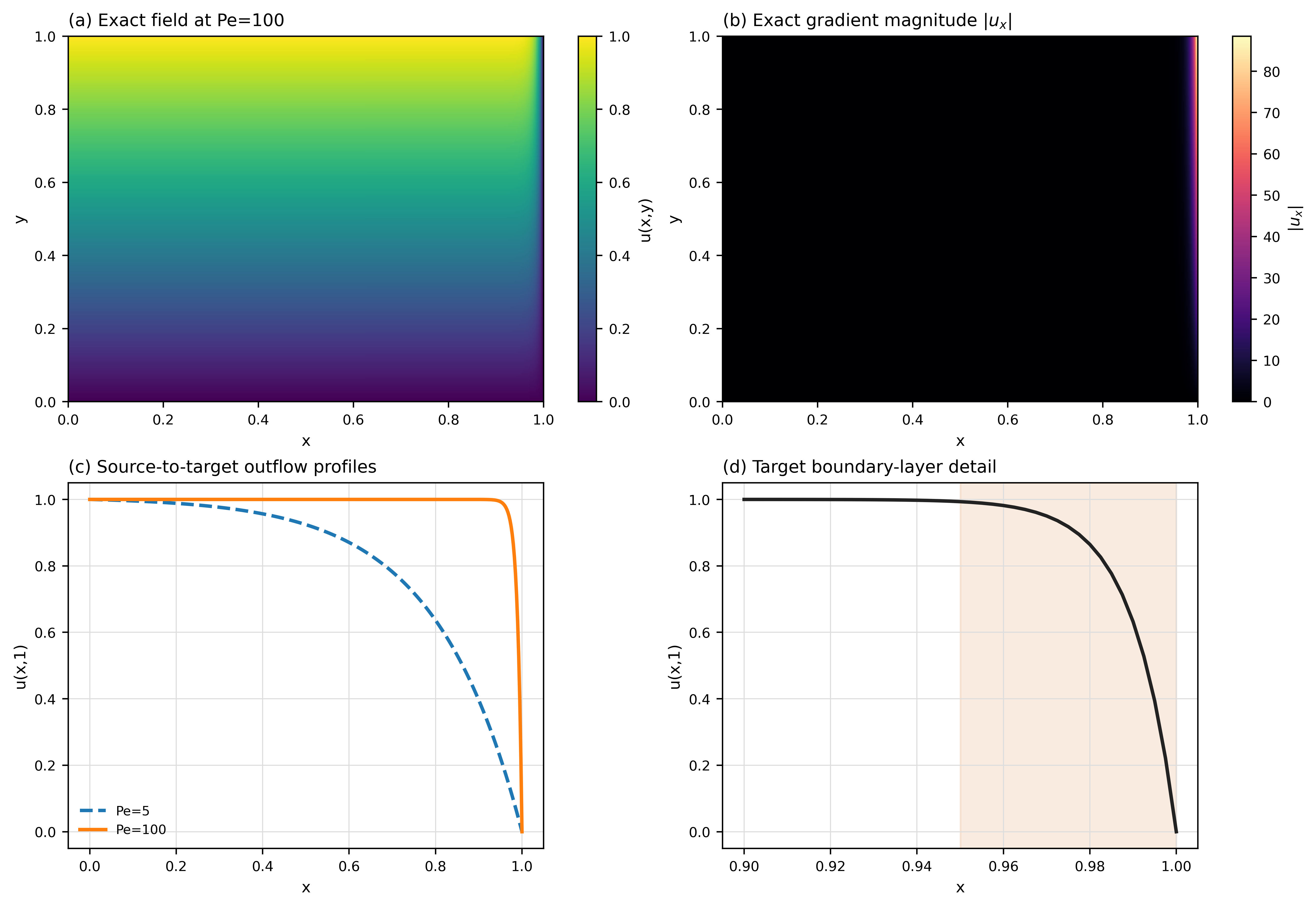}
  \caption{\revision{Structure of the high-P\'{e}clet inflow--outflow benchmark. The exact field, $|u_x|$, and outflow profiles show the source--target change from $Pe=5$ to $Pe=100$ and the boundary layer near $x=1$.}}
  \label{fig:pe_boundary_structure}
\end{figure}

\begin{table}[t]
\centering
\caption{\revision{High-P\'{e}clet boundary-layer inverse-problem results.}}
\label{tab:pe_results}
\scriptsize
\begin{tabularx}{\columnwidth}{@{}lYY@{}}
\toprule
Method & Global field error (\%) & Parameter error (\%) \\
\midrule
Standard PINNs & $(4.976\pm15.740)\!\times\!10^{11}$ & $(1.251\pm3.955)\!\times\!10^{13}$ \\
\revision{Full} & $0.1291\pm0.0337$ & $0.3786\pm0.2193$ \\
Lightweight FT & $0.3959\pm0.1536$ & $6.5651\pm3.5718$ \\
Partial Transfer & $0.0931\pm0.0386$ & $0.4467\pm0.2316$ \\
TL-gPINN & $\mathbf{0.0860\pm0.0195}$ & $0.2510\pm0.1259$ \\
BitFit & $13.0062\pm0.8695$ & $947.1331\pm221.1255$ \\
L2-SP & $0.1505\pm0.0386$ & $0.4117\pm0.3735$ \\
SFP & $10.4402\pm21.8107$ & $(4.218\pm13.340)\!\times\!10^4$ \\
\textbf{TGSR-PINN} & $0.0946\pm0.0216$ & $\mathbf{0.1260\pm0.1148}$ \\
\bottomrule
\end{tabularx}
\end{table}

\begin{revisionblock}
TGSR-PINN has the lowest mean parameter error among the compared methods. Relative to Full, its parameter, global-field, boundary-layer, and gradient errors decrease by 66.7\%, 26.8\%, 24.6\%, and 26.7\%, respectively. TL-gPINN has a slightly lower mean global field error (0.0860\% versus 0.0946\%), but its mean parameter error is higher (0.2510\% versus 0.1260\%). Thus, TGSR-PINN improves parameter recovery while keeping the mean field error below 0.1\%.
\end{revisionblock}

\begin{table}[t]
\centering
\caption{\revision{Boundary-layer and gradient-error diagnostics.}}
\label{tab:pe_local_metrics}
\small
\begin{tabularx}{\columnwidth}{@{}lYYYY@{}}
\toprule
Method & Param. (\%) & Global (\%) & Layer (\%) & Gradient \\
\midrule
\revision{Full} & 0.3786 & 0.1291 & 0.3369 & 0.010462 \\
L2-SP & 0.4117 & 0.1505 & 0.3659 & 0.011453 \\
SFP & 42184.8503 & 10.4402 & 11.6929 & 0.206865 \\
\textbf{TGSR-PINN} & \textbf{0.1260} & \textbf{0.0946} & \textbf{0.2540} & \textbf{0.007671} \\
\bottomrule
\end{tabularx}
\end{table}

\begin{figure*}[t]
  \centering
  \includegraphics[width=.90\textwidth]{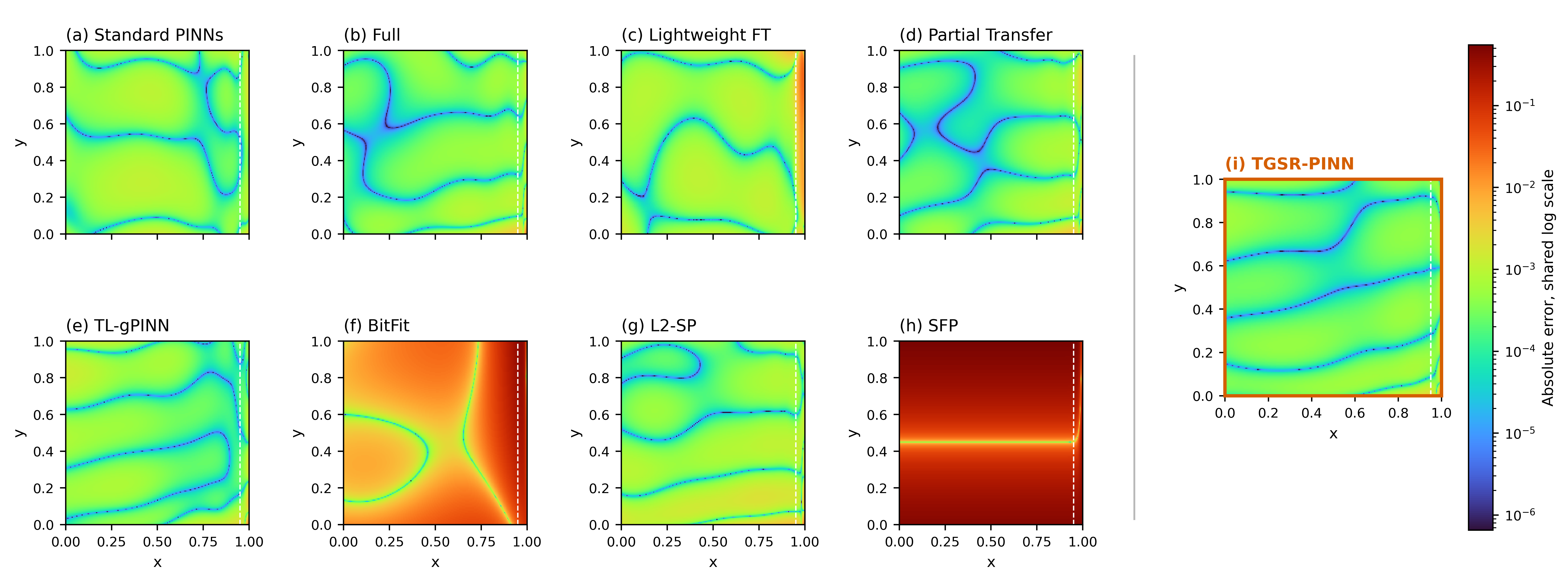}
  \caption{\revision{Absolute-error maps for all nine methods in a representative paired repetition. The baselines are arranged on the left, and TGSR-PINN is shown separately on the right. All panels use the same logarithmic scale.}}
  \label{fig:pe_all_errors}
\end{figure*}

\begin{figure}[tbp]
  \centering
  \includegraphics[width=.84\columnwidth]{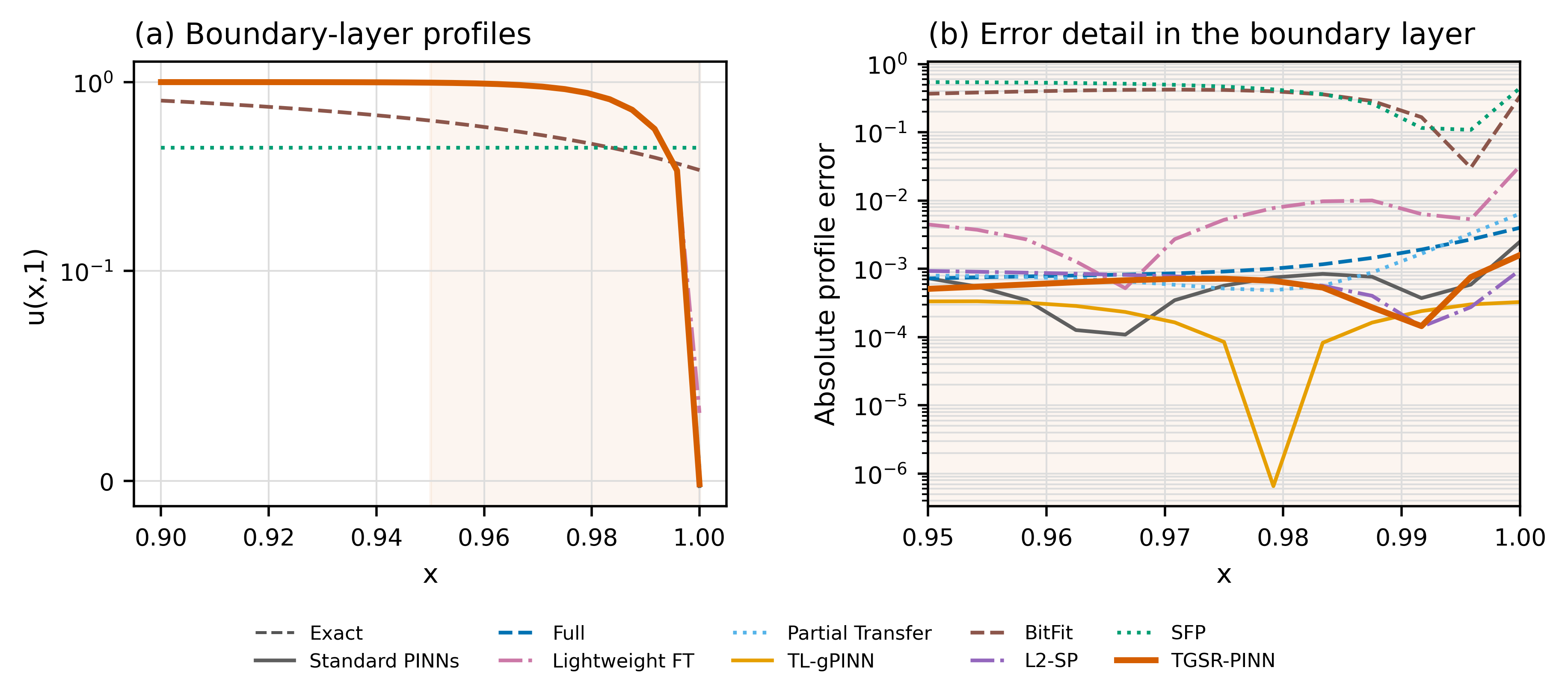}
  \caption{\revision{Outflow-layer profiles at $y=1$ for all nine methods in the same representative paired repetition. Panel (a) compares the predicted profiles with the exact solution. Panel (b) shows the absolute errors over $x\geq0.95$ on a logarithmic scale.}}
  \label{fig:pe_profiles}
\end{figure}

\subsubsection*{Paired statistical tests over ten random-seed repetitions}

\begin{table}[tbp]
\centering
\caption{\revision{Paired Wilcoxon comparisons on the high-P\'{e}clet boundary-layer task.}}
\label{tab:pe_wilcoxon}
\scriptsize
\begin{tabularx}{\columnwidth}{@{}lYYYY@{}}
\toprule
& \multicolumn{2}{c}{Parameter error} & \multicolumn{2}{c}{Field error} \\
\cmidrule(lr){2-3}\cmidrule(lr){4-5}
Baseline & Wins & $p$ & Wins & $p$ \\
\midrule
Standard PINNs & $9/10$ & $\mathbf{0.00977}$ & $8/10$ & $\mathbf{0.03711}$ \\
\revision{Full} & $10/10$ & $\mathbf{0.00195}$ & $9/10$ & $\mathbf{0.03711}$ \\
Lightweight FT & $10/10$ & $\mathbf{0.00195}$ & $10/10$ & $\mathbf{0.00195}$ \\
Partial Transfer & $9/10$ & $\mathbf{0.00586}$ & $5/10$ & $0.92188$ \\
TL-gPINN & $10/10$ & $\mathbf{0.00195}$ & $5/10$ & $0.43164$ \\
BitFit & $10/10$ & $\mathbf{0.00195}$ & $10/10$ & $\mathbf{0.00195}$ \\
L2-SP & $8/10$ & $\mathbf{0.01367}$ & $9/10$ & $\mathbf{0.01367}$ \\
SFP & $9/10$ & $\mathbf{0.00977}$ & $7/10$ & $0.10547$ \\
\bottomrule
\end{tabularx}
\end{table}

\begin{revisionblock}
TGSR-PINN achieves lower parameter error than every baseline in at least 8 of the 10 paired repetitions, and all eight parameter-error comparisons yield $p<0.05$. The field-error comparisons are also significant against Standard PINNs, Full, Lightweight FT, BitFit, and L2-SP. Against Partial Transfer, TL-gPINN, and SFP, the field-error differences are not statistically significant. The paired analysis therefore shows a consistent parameter-recovery advantage while the field error remains at the level of the stable transfer baselines.
\end{revisionblock}

\begin{figure}[t]
  \centering
  \includegraphics[width=.72\columnwidth]{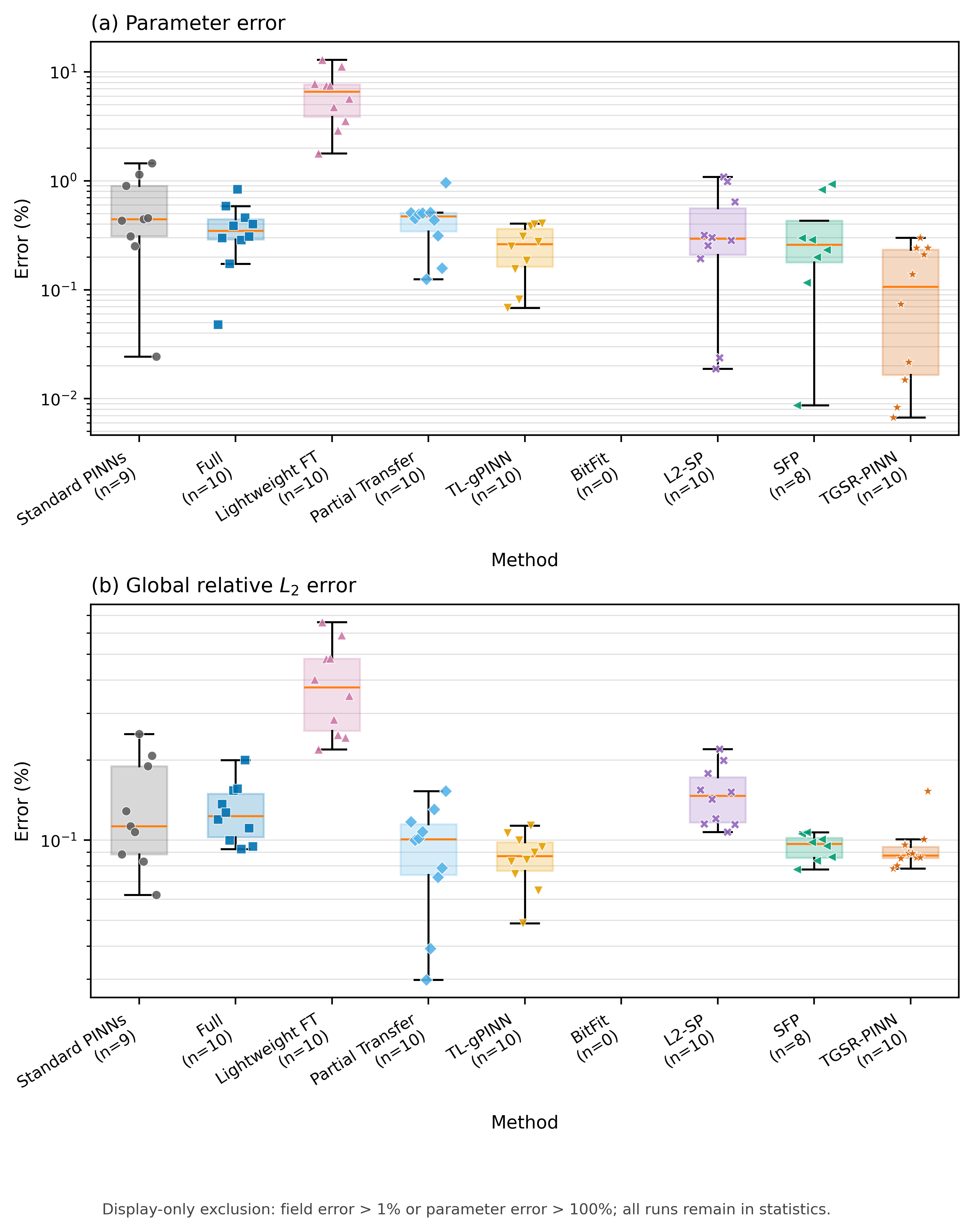}
  \caption{\revision{Parameter- and global-field-error distributions within the common visualization range. The retained sample count is shown below each method.}}
  \label{fig:pe_error_distributions}
\end{figure}

One Standard-PINN repetition exhibits numerical divergence. Across the remaining nine repetitions, the mean (median) field error is 0.1368\% (0.1128\%), and the mean (median) parameter error is 0.6014\% (0.4457\%). All ten BitFit repetitions fall outside the visualization range, with field errors of 12.438--15.341\% and parameter errors of 783.242--1543.888\%, showing that bias-only adaptation is insufficient for this source--target shift. Against the numerically stable \revision{Full} and TL-gPINN baselines, TGSR-PINN achieves lower parameter error in all ten paired repetitions for both comparisons ($p=0.00195$).

\begin{table}[b]
\centering
\caption{\revision{Visualization-range exclusions for Figure~\ref{fig:pe_error_distributions}.}}
\label{tab:pe_failed_runs}
\scriptsize
\setlength{\tabcolsep}{3pt}
\begin{tabularx}{\columnwidth}{@{}lcYY@{}}
\toprule
Method & Failed & Field error (\%) & Parameter error (\%) \\
\midrule
Standard PINNs & 1/10 & $4.976\times10^{12}$ & $1.251\times10^{14}$ \\
BitFit & 10/10 & $12.438$--$15.341$ & $783.242$--$1543.888$ \\
SFP & 2/10 & $51.750,\ 51.896$ & $100.000,\ 421745.589$ \\
\bottomrule
\end{tabularx}
\end{table}
\end{revisionblock}

\begin{figure*}[t]
  \centering
  \includegraphics[width=.90\textwidth]{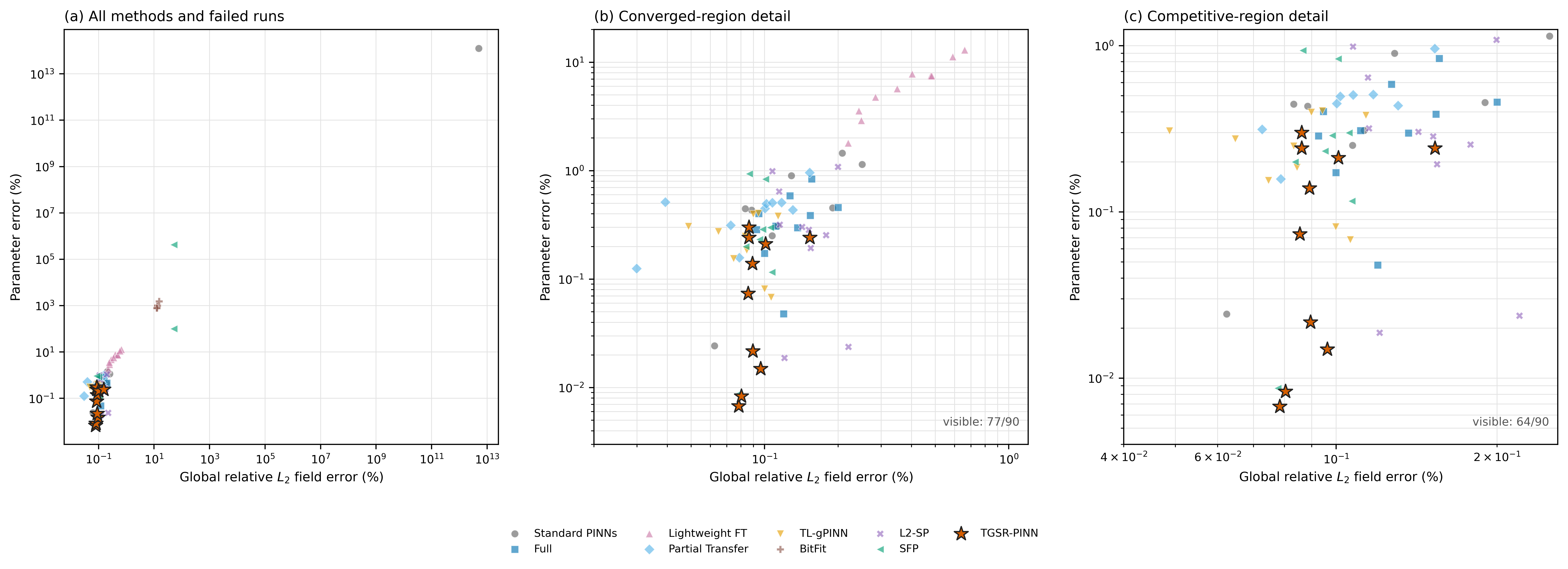}
  \caption{\revision{Joint global-field and parameter errors for all nine methods. TGSR-PINN is highlighted by larger orange star markers. The full-range panel retains every run, and the detail panels change only the axis limits.}}
  \label{fig:field_parameter_relation}
\end{figure*}

\subsection{Cross-PDE-Family Transfer: Allen--Cahn \texorpdfstring{$\to$}{to} Burgers Inverse Problem}
\label{subsec:cross_pde}

\revision{The high-P\'{e}clet benchmark primarily examines within-family transfer from a lower-P\'{e}clet advection--diffusion source task ($Pe=5$) to a strongly advection-dominated boundary-layer target task ($Pe=100$). To further examine transfer across governing-equation families, we construct an Allen--Cahn--Burgers cross-PDE-family transfer task. This task requires the model to transfer from a reaction--diffusion-dominated source model to a nonlinear advection and viscous diffusion target inverse problem, making it more likely to expose source representation bias and target parameter compensation issues.}

The source task is a 1D Allen--Cahn inverse problem:
\begin{equation}
  u_{t} = \revision{\epsilon_{\mathrm{AC}}}^{2}u_{xx} + u - u^{3},\quad\text{infer }\revision{\epsilon_{\mathrm{AC}}}.
\end{equation}
The target task is a 1D Burgers inverse problem:
\begin{equation}
  u_{t} + uu_{x} = \nu u_{xx},\quad\text{infer }\nu.
\end{equation}
The Allen--Cahn source task uses $\epsilon_{\mathrm{AC}}=0.85$ as the initial value and $\epsilon_{\mathrm{AC}}=0.8$ as the true value, with 1\% observation noise. The Burgers target task uses $\nu=0.05$ as the initial value and $\nu=0.01$ as the true value, with 3\% observation noise. \revision{L2-SP and SFP are included as regularization and soft-pruning baselines, respectively.}

\begin{revisionblock}
\begin{table}[htbp]
\centering
\caption{\revision{Allen--Cahn $\to$ Burgers cross-PDE transfer results.}}
\label{tab:cross_results}
\scriptsize
\begin{tabularx}{\columnwidth}{@{}lcYY@{}}
\toprule
Method & $n$ & $L_2$ error (\%) & $\nu$ error (\%) \\
\midrule
PINNs              & 10 & $0.3102 \pm 0.0894$ & $0.8230 \pm 0.8182$ \\
\revision{Full}    & 10 & $0.2706 \pm 0.0529$ & $0.7618 \pm 0.4027$ \\
Lightweight FT     & 10 & $0.8422 \pm 0.4537$ & $13.3479 \pm 30.6113$ \\
Partial Transfer   & 10 & $0.2982 \pm 0.0599$ & $0.9098 \pm 0.6309$ \\
TL-gPINN           & 10 & $0.2943 \pm 0.0567$ & $0.6501 \pm 0.4584$ \\
BitFit             & 10 & $68.5002 \pm 15.9476$ & $5348.6681 \pm 14431.0975$ \\
L2-SP              & 10 & $0.2739 \pm 0.0658$ & $0.6117 \pm 0.3540$ \\
SFP                & 10 & $0.2968 \pm 0.0528$ & $0.4263 \pm 0.3058$ \\
\textbf{TGSR-PINN} & 10 & $\mathbf{0.2644 \pm 0.0593}$ & $\mathbf{0.3071 \pm 0.2140}$ \\
\bottomrule
\end{tabularx}
\end{table}

Table~\ref{tab:cross_results} shows that TGSR-PINN outperforms the compared methods in both mean field error and mean viscosity-parameter error over the ten paired repetitions. Relative to \revision{Full}, its mean $\nu$ error decreases from 0.7618\% to 0.3071\%; relative to SFP, it decreases from 0.4263\% to 0.3071\%. The result supports the intended role of target-guided representation correction under a substantial source--target mechanism mismatch. The very large BitFit errors indicate that bias-only adaptation does not provide sufficient representation flexibility for this cross-family transfer setting.
\revision{The paired tests in Table~\ref{tab:secondary_task_wilcoxon} show that TGSR-PINN has lower parameter error than Full in 9/10 comparisons ($p=0.01367$). Its mean parameter error is also lower than those of SFP and TL-gPINN, with paired $p=0.49219$ and $p=0.10547$, respectively.}
\end{revisionblock}

\subsection{2D Reaction--Diffusion Inverse Problem: 5\% Noise}
\label{subsec:reaction_diffusion}

Following the high-P\'{e}clet main experiment and cross-PDE-family transfer, we further examine a dual-parameter inverse problem with milder source--target differences but noisy observation data. The 2D reaction--diffusion target task is:
\begin{equation}
  \begin{aligned}
  u_{t} - \alpha(u_{xx} + u_{yy}) + \rho u^{2}
  &= Q_{T}(x,y,t),\\
  \text{infer}\quad
  &\alpha,\rho.
  \end{aligned}
\end{equation}
This task transfers from a 2D diffusion source task to a target task with a reaction term, inferring diffusion coefficient $\alpha$ and reaction coefficient $\rho$. \revision{Here $Q_T(x,y,t)$ denotes the target source term. The source--target difference is milder than in the high-P\'{e}clet inflow--outflow benchmark because both tasks retain diffusion dynamics, while the target task introduces an additional nonlinear reaction term. A 5\% noise level is applied only to the target observations.}

\begin{revisionblock}
\begin{table}[htbp]
\centering
\caption{\revision{Results for the 5\%-noise 2D reaction--diffusion inverse problem.}}
\label{tab:rd_results}
\scriptsize
\begin{tabularx}{\columnwidth}{@{}lcYY@{}}
\toprule
Method & $n$ & $L_2$ error (\%) & Param error (\%) \\
\midrule
PINNs              & 10 & $0.2153 \pm 0.0240$ & $0.8217 \pm 0.6406$ \\
\revision{Full}    & 10 & $0.2235 \pm 0.0200$ & $1.3999 \pm 0.7983$ \\
Lightweight FT     & 10 & $0.2380 \pm 0.0302$ & $1.1680 \pm 0.8982$ \\
Partial Transfer   & 10 & $0.2410 \pm 0.0328$ & $1.6333 \pm 1.2408$ \\
TL-gPINN           & 10 & $0.2193 \pm 0.0182$ & $1.2281 \pm 0.9593$ \\
BitFit             & 10 & $0.2711 \pm 0.0542$ & $1.1293 \pm 0.7429$ \\
L2-SP              & 10 & $0.2256 \pm 0.0252$ & $1.4148 \pm 0.9952$ \\
SFP                & 10 & $0.2239 \pm 0.0235$ & $1.3907 \pm 0.7604$ \\
\textbf{TGSR-PINN} & 10 & $\mathbf{0.2118 \pm 0.0223}$ & $\mathbf{0.6581 \pm 0.5714}$ \\
\bottomrule
\end{tabularx}
\end{table}
\end{revisionblock}

\begin{revisionblock}
\begin{table}[tbp]
\centering
\caption{Paired Wilcoxon comparisons for the Cross-PDE and reaction--diffusion tasks.}
\label{tab:secondary_task_wilcoxon}
\scriptsize
\setlength{\tabcolsep}{3pt}
\begin{tabularx}{\columnwidth}{@{}llYY@{}}
\toprule
Task & Baseline & Parameter wins / $p$ & Field wins / $p$ \\
\midrule
Cross-PDE & Standard PINNs & $8/10$ / $\mathbf{0.04883}$ & $6/10$ / $0.37500$ \\
          & Full          & $9/10$ / $\mathbf{0.01367}$ & $5/10$ / $0.62500$ \\
          & L2-SP         & $9/10$ / $\mathbf{0.01953}$ & $5/10$ / $0.84570$ \\
          & SFP           & $6/10$ / $0.49219$ & $7/10$ / $0.19336$ \\
          & TL-gPINN      & $7/10$ / $0.10547$ & $8/10$ / $0.10547$ \\
\midrule
Reaction--diffusion & Standard PINNs & $6/10$ / $0.49219$ & $7/10$ / $0.43164$ \\
          & Full          & $8/10$ / $\mathbf{0.00977}$ & $8/10$ / $\mathbf{0.01953}$ \\
          & L2-SP         & $9/10$ / $\mathbf{0.01367}$ & $9/10$ / $\mathbf{0.01367}$ \\
          & SFP           & $10/10$ / $\mathbf{0.00195}$ & $8/10$ / $0.06445$ \\
          & TL-gPINN      & $9/10$ / $\mathbf{0.04883}$ & $7/10$ / $0.10547$ \\
\bottomrule
\end{tabularx}
\end{table}
\end{revisionblock}

\begin{figure*}[tbp]
  \centering
  \includegraphics[width=.94\textwidth]{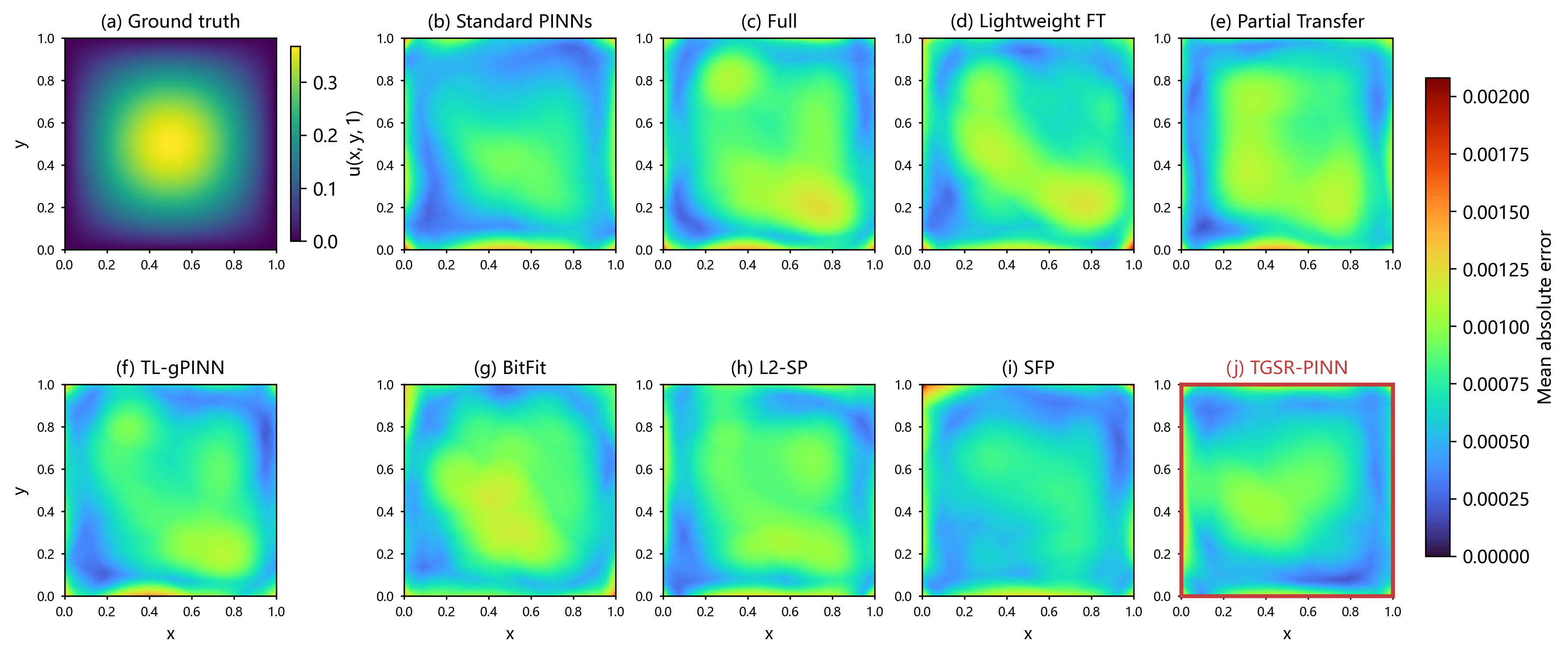}
  \caption{\revision{Ground-truth field and pointwise mean absolute-error maps at $t=1$ for the 5\%-noise reaction--diffusion task. All methods use the same evaluation grid and linear color scale.}}
  \label{fig:rd_field}
\end{figure*}

\begin{revisionblock}
Table~\ref{tab:rd_results} shows that TGSR-PINN outperforms the compared methods in mean field error and mean parameter error in the 5\%-noise setting. Its mean parameter error is 0.6581\%, compared with 0.8217\% for standard PINNs and 1.3999\% for \revision{Full}. The field-error differences are smaller than the parameter-error differences, again illustrating why field reconstruction alone is insufficient for evaluating inverse transfer. Figure~\ref{fig:rd_field} provides the corresponding pointwise mean absolute-error maps.

\revision{Table~\ref{tab:secondary_task_wilcoxon} further shows that TGSR-PINN has significantly lower parameter error than Full, L2-SP, SFP, and TL-gPINN in the paired comparisons. Its field error is also significantly lower than Full and L2-SP, while the comparison with standard PINNs does not reach $p<0.05$ for either metric.}

Figure~\ref{fig:rd_field} shows that the general reaction--diffusion field structure can be reconstructed under 5\% observation noise, while TGSR-PINN produces a relatively lower absolute error pattern. This noisy dual-parameter task therefore complements the boundary-layer and cross-PDE experiments with evidence under observation corruption and milder source--target mismatch.
\end{revisionblock}

\begin{revisionblock}
\subsection{Smooth Manufactured-Solution Advection--Diffusion Inverse Problem}
\label{subsec:smooth_pe_diagnostic}

We additionally evaluate a time-dependent two-dimensional manufactured-solution advection--diffusion inverse problem. The source task contains diffusion only, whereas the target task is
\begin{equation}
  u_t-\alpha(u_{xx}+u_{yy})+v_xu_x+v_yu_y=Q_T(x,y,t),
  \qquad \text{infer }\alpha,v_x,v_y,
  \label{eq:smooth_pe_target}
\end{equation}
with the reference field $u=e^{-t}\sin(\pi x)\sin(\pi y)$, $\alpha=0.001$, $v_x=2$, and $v_y=1$. The resulting nominal P\'{e}clet number is \revision{$Pe=U_cL_c/\alpha\approx2.24\times10^3$}, where \revision{$U_c$ and $L_c$ denote the characteristic velocity and length used for this estimate}. The manufactured forcing $Q_T$ yields a globally smooth reference field with compatible boundary data, so this experiment evaluates optimization behavior and coupled parameter recovery under a controlled smooth solution.

\begin{table*}[t]
\centering
\caption{\revision{Smooth manufactured-solution advection--diffusion inverse-problem results.}}
\label{tab:smooth_pe_results}
\small
\begin{tabularx}{\textwidth}{@{}lYYYY@{}}
\toprule
Method & $L_2$ error (\%) & Param. error (\%) & Best $\alpha$ error (\%) & Time (s) \\
\midrule
PINNs              & $0.134\pm0.038$ & $14.825\pm6.021$ & 14.63 & 252 \\
\revision{Full}    & $0.088\pm0.018$ & $6.582\pm3.314$  & 10.68 & 235 \\
Lightweight FT     & $0.087\pm0.011$ & $4.825\pm2.384$  & 4.06  & 232 \\
Partial Transfer   & $0.129\pm0.027$ & $13.906\pm4.332$ & 24.25 & 242 \\
TL-gPINN           & $0.091\pm0.022$ & $7.430\pm3.320$  & 11.63 & 552 \\
BitFit             & $0.184\pm0.035$ & $7.137\pm4.754$  & 4.09  & 160 \\
\textbf{TGSR-PINN} & $\mathbf{0.080\pm0.016}$ & $\mathbf{4.426\pm3.019}$ & \textbf{0.38} & 266 \\
\bottomrule
\end{tabularx}
\end{table*}

Table~\ref{tab:smooth_pe_results} shows that TGSR-PINN produces lower mean field error, mean parameter error, and best observed $\alpha$ error than the compared methods, supporting target-side representation correction for coupled parameter recovery on the smooth solution manifold.
\end{revisionblock}

\subsection{Mechanism Ablations and Sensitivity Analysis on the Smooth Advection--Diffusion Task}
\label{subsec:ablation}

\begin{revisionblock}
The smooth manufactured-solution task provides a controlled setting for analyzing the internal components of TGSR-PINN. The target-adaptation, random-reset, scoring-batch, data-sparsity, and adaptation-length studies examine local mechanism trends. The counterfactual, layer-protection, and mapping studies use ten paired repetitions, while the GMM/rank comparison and the $\alpha_s$ sweep evaluate distribution routing and score-fusion sensitivity.
\end{revisionblock}

\subsubsection{Mechanism ablation: target short adaptation, random reset, and selective soft decay}

In the mechanism ablation experiment, we compare three settings: Target-Adaptation Only (only target short adaptation, no neuron target scoring or subsequent parameter reorganization), Random Reset (retains neuron target scoring and weak-adaptation signal estimation but replaces selective soft decay with random initialization-style resetting), and TGSR-PINN (complete method).

\begin{table}[htbp]
\centering
\caption{\revision{Mechanism ablation on the smooth advection--diffusion task.}}
\label{tab:ablation_early}
\small
\begin{tabularx}{\columnwidth}{@{}lYY@{}}
\toprule
Method & $L_2$ error (\%) & Param error (\%) \\
\midrule
Target-Adaptation Only & $0.144 \pm 0.010$ & $19.577 \pm 1.689$ \\
Random Reset           & $0.168 \pm 0.009$ & $19.457 \pm 0.979$ \\
\textbf{TGSR-PINN}     & $\mathbf{0.069 \pm 0.006}$ & $\mathbf{2.555 \pm 2.024}$ \\
\bottomrule
\end{tabularx}
\end{table}

\begin{revisionblock}
Target-Adaptation Only leaves the mean parameter error at 19.577\%, and Random Reset gives a similar error of 19.457\%. TGSR-PINN reduces the mean parameter error to 2.555\% and the relative $L_2$ field error to 0.069\%. In this early diagnostic, short adaptation alone and random resetting do not reproduce the parameter recovery of the full method. The paired ablations below examine neuron scoring, score--neuron correspondence, layer protection, and the selective soft-decay mapping separately.
\end{revisionblock}

To further test whether the full method's benefit comes from neuron target scoring and neuron correspondence rather than general weight perturbation, we run paired counterfactual ablations over 10 paired repetitions under the same source model, target task, noise level, target data settings, and optimization budget. Results are shown in Table~\ref{tab:ablation_counterfactual}.

\begin{table}[htbp]
\centering
\caption{\revision{Paired counterfactual ablation.}}
\label{tab:ablation_counterfactual}
\small
\resizebox{\columnwidth}{!}{%
\begin{tabular}{lccccc}
\toprule
Method & $n$ & $L_2$ (\%) & Param (\%) & $\alpha$ (\%) & Wilcoxon \\
\midrule
\textbf{TGSR-PINN}     & 10 & $\mathbf{0.074\pm0.015}$ & $\mathbf{3.782\pm3.068}$ & $\mathbf{11.312\pm9.205}$ & \revision{Reference} \\
Random Soft Decay      & 10 & $0.084\pm0.012$ & $5.216\pm2.377$ & $15.613\pm7.122$ & $p=0.037$ \\
Score-Shuffled         & 10 & $0.084\pm0.023$ & $5.907\pm3.574$ & $17.688\pm10.727$ & $p=0.014$ \\
\bottomrule
\end{tabular}}
\end{table}

\begin{revisionblock}
TGSR-PINN has a mean parameter error of 3.782\% $\pm$ 3.068\%, compared with 5.216\% $\pm$ 2.377\% for Random Soft Decay and 5.907\% $\pm$ 3.574\% for Score-Shuffled TGSR. TGSR-PINN wins 8/10 paired comparisons against each counterfactual. The median paired increases are $+1.065$ and $+2.278$ percentage points, with two-sided Wilcoxon $p=0.037$ and $p=0.014$, respectively. The two controls show that target-informed neuron selection and preservation of the score--neuron correspondence both contribute to parameter recovery.
\end{revisionblock}

\subsubsection{Scoring batch count ablation}

\begin{revisionblock}
This subsection examines how neuron target scoring responds to the number of scoring batches and the amount of observation data. These early diagnostics are used to inspect the mechanism under sampling changes; the multi-repetition experiments provide the main robustness evidence.
\end{revisionblock}

\begin{table}[htbp]
\centering
\caption{\revision{Scoring batch count ablation.}}
\label{tab:batch_ablation}
\small
\begin{tabularx}{\columnwidth}{@{}YYY@{}}
\toprule
$K$ & $L_2$ error (\%) & Param error (\%) \\
\midrule
1 & 0.0693 & 2.662 \\
3 & 0.0693 & 2.555 \\
\bottomrule
\end{tabularx}
\end{table}

Table~\ref{tab:batch_ablation} shows that increasing the scoring batch count from $K=1$ to $K=3$ leaves the $L_2$ error essentially unchanged while the parameter error decreases slightly from 2.662\% to 2.555\%. This suggests that multi-batch scoring may reduce sensitivity to a single scoring batch and yield a modest improvement in this diagnostic.

\begin{table}[htbp]
\centering
\caption{\revision{Data sparsity stress test.}}
\label{tab:sparsity}
\small
\begin{tabularx}{\columnwidth}{@{}lYY@{}}
\toprule
Method & $N_d=1000$ & $N_d=50$ \\
\midrule
PINNs         & 19.1\% / 0.09\% & 30.6\% / 0.12\% \\
\revision{Full} & 19.7\% / 0.10\% & 16.8\% / 0.11\% \\
\textbf{TGSR-PINN} & \textbf{2.3\% / 0.09\%} & \textbf{11.4\% / 0.10\%} \\
\bottomrule
\end{tabularx}
\end{table}

Table~\ref{tab:sparsity} presents a single-run data sparsity stress test. When the observation count drops from $N_d=1000$ to $N_d=50$, TGSR-PINN's $\alpha$ error increases from 2.3\% to 11.4\%, indicating that extreme sparsity weakens parameter recovery; however, compared to PINNs (30.6\%) and \revision{Full} (16.8\%), TGSR-PINN still maintains lower error in this stress test.

\begin{table}[htbp]
\centering
\caption{\revision{Target short adaptation length ablation.}}
\label{tab:ead_length}
\small
\begin{tabularx}{\columnwidth}{@{}YYYY@{}}
\toprule
$E_{ad}$ & Rep A $\alpha$ error & Rep B $\alpha$ error & Fraction of main \\
\midrule
10  & 14.6\% & 12.5\% & 6.7\% \\
20  & 15.7\% & 6.3\%  & 13.3\% \\
\textbf{30 (default)} & \textbf{9.2\%} & \textbf{6.2\%} & \textbf{20\%} \\
60  & 11.4\% & 6.9\%  & 40\% \\
\bottomrule
\end{tabularx}
\end{table}

\begin{revisionblock}
Table~\ref{tab:ead_length} reports a preliminary sensitivity check of target short adaptation length over two random repetitions. $E_{ad}=30$, corresponding to 20\% of main training, gives the lowest or a comparable $\alpha$ error in both repetitions. We therefore use it as the fixed setting in the subsequent diagnostics. Task-adaptive selection of $E_{ad}$ is discussed in Section~\ref{subsec:cost}.
\end{revisionblock}

\subsubsection{Scoring component ablation: \texorpdfstring{$\alpha_s$}{alpha_s} parameter sweep}

\revision{To examine the sensitivity of neuron target scoring to the relative contributions of Taylor sensitivity and pre-activation variance, we sweep $\alpha_s$ from 0 to 1 in increments of 0.1 on the smooth manufactured-solution advection--diffusion task, using five random repetitions per setting.}

\begin{table}[htbp]
\centering
\caption{\revision{$\alpha_s$ fusion-weight sweep.}}
\label{tab:alpha_s}
\scriptsize
\begin{tabularx}{\columnwidth}{@{}YYYYY@{}}
\toprule
$\alpha_s$ & $\alpha$ mean & Best & Worst & All-param mean \\
\midrule
0.0 & 17.9\% & 1.6\%  & 34.4\% & 6.0\% \\
0.1 & 14.4\% & 5.3\%  & 30.3\% & 4.8\% \\
0.2 & 29.2\% & 3.2\%  & 78.2\% & 9.7\% \\
0.3 & 15.9\% & \textbf{0.4\%} & 40.7\% & 5.3\% \\
0.4 & 17.4\% & 2.4\%  & 29.1\% & 5.8\% \\
0.5 (default) & 13.8\% & 2.3\% & 38.1\% & 4.6\% \\
0.6 & 16.2\% & 1.3\%  & 28.0\% & 5.4\% \\
0.7 & 15.2\% & 7.8\%  & 35.7\% & 5.1\% \\
0.8 & \textbf{13.2\%} & 1.6\% & 38.5\% & \textbf{4.4\%} \\
0.9 & 18.1\% & 8.3\%  & 40.5\% & 6.1\% \\
1.0 & 14.6\% & 1.9\% & 33.3\% & 4.9\% \\
\bottomrule
\end{tabularx}
\end{table}

\begin{figure}[tbp]
  \centering
  \includegraphics[width=.94\columnwidth]{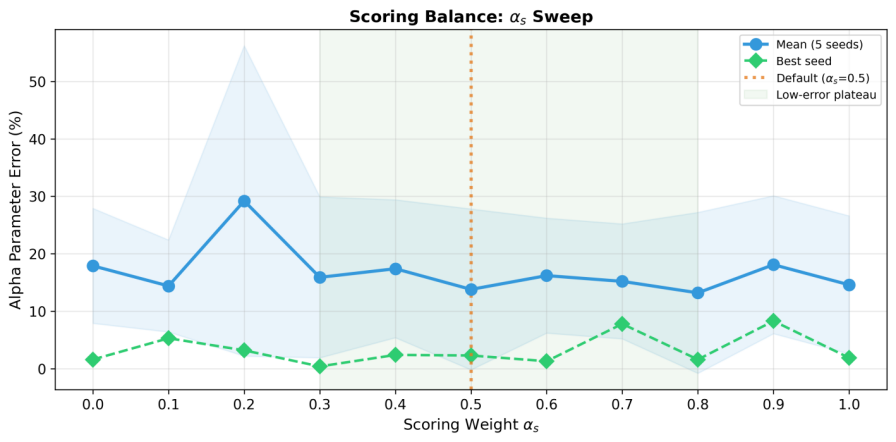}
  \caption{\revision{Sensitivity of parameter recovery to the fusion weight $\alpha_s$. The response is non-monotonic, and $\alpha_s=0.8$ yields the lowest mean error over the tested values.}}
  \label{fig:alpha_sweep}
\end{figure}

\begin{revisionblock}
The $\alpha_s=0.5$ joint-scoring setting achieves a mean $\alpha$ error of 13.8\% in this mechanism sweep over five random repetitions. We retain $\alpha_s=0.5$ as a balanced fixed default. Figure~\ref{fig:alpha_sweep} reports the complete fusion-weight sensitivity curve.

We also calibrate the score against a 10\% single-neuron pre-activation gate over two fixed repetitions. This offline diagnostic uses 30 probes per repetition across six hidden layers without updating the model or physical parameters. Across the pooled 60 probes, the target score has Spearman correlation $\rho_{\mathrm{S}}=0.7466$ with the observed target-loss change. The parameter-mixed sensitivity diagnostic has $\rho_{\mathrm{S}}=0.9921$ with the observed physical-parameter-gradient change. At the evaluated target states, both correlations are consistent with the corresponding local perturbation responses.
\end{revisionblock}

\subsubsection{GMM and rank fallback ablation}

\begin{revisionblock}
TGSR-PINN employs a dual-path, distribution-adaptive design for weak-adaptation signal estimation. Its purpose is to avoid assigning the same numerical cutoff or the same intervention proportion to layers with different score scales and distribution shapes. When the model-selection and \revision{non-majority low-mean-component} conditions are satisfied, the GMM branch converts layer-specific posterior responsibilities into continuous weak-adaptation signals. Otherwise, rank fallback produces bounded monotone signals without declaring a predetermined fraction of neurons unsuitable. GMM therefore provides conditional data-driven grouping, while rank fallback keeps the signal estimator well defined across layers that do not exhibit reliable mixture structure.
\end{revisionblock}

\begin{table}[htbp]
\centering
\caption{\revision{GMM and rank fallback ablation.}}
\label{tab:gmm_ablation}
\small
\resizebox{\columnwidth}{!}{%
\begin{tabular}{lccccc}
\toprule
Method & $n$ & $L_2$ (\%) & Param (\%) & $\alpha$ (\%) & Wilcoxon \\
\midrule
TGSR (GMM+rank) & 5 & $0.071\pm0.007$ & $2.944\pm2.348$ & $8.805\pm7.044$ & \revision{Reference} \\
TGSR (rank only) & 5 & $0.069\pm0.007$ & $3.090\pm2.380$ & $9.239\pm7.140$ & $p=0.593$ \\
\bottomrule
\end{tabular}}
\end{table}

\begin{revisionblock}
The rank-only branch produces field and parameter errors similar to those of the dual-path estimator in this diagnostic; the paired difference is not statistically significant ($p=0.593$). This is consistent with using rank-based signaling when mixture evidence is insufficient. Branch activation records make the layer-wise routing decisions directly inspectable.

We further evaluate all three branch choices over 20 random Allen--Cahn to Burgers repetitions, where the GMM condition activates in 18/20 repetitions and 34/120 hidden layers. Table~\ref{tab:gmm_cross_ablation} shows that GMM-only obtains the lowest mean parameter error, whereas GMM+rank obtains the lowest mean field error. The two metrics therefore exhibit a parameter--field tradeoff rather than a uniform advantage of one branch. We retain GMM for conditional grouping when the score distribution supports a low-score component and use rank fallback when that evidence is absent. This routing avoids a fixed score cutoff and a fixed neuron-intervention ratio across layers.

\begin{table}[htbp]
\centering
\caption{\revision{Cross-PDE GMM/rank branch comparison.}}
\label{tab:gmm_cross_ablation}
\small
\begin{tabularx}{\columnwidth}{@{}lYY@{}}
\toprule
Branch & Parameter error (\%) & $L_2$ error (\%) \\
\midrule
GMM+rank & $1.4383\pm1.5100$ & $\mathbf{0.3011\pm0.1079}$ \\
Rank-only & $1.2735\pm1.1050$ & $0.3040\pm0.1033$ \\
GMM-only & $\mathbf{1.0810\pm1.0718}$ & $0.3056\pm0.0934$ \\
\bottomrule
\end{tabularx}
\end{table}

\begin{table}[tbp]
\centering
\caption{Sensitivity of the GMM/rank switching rule to network depth and empirical thresholds.}
\label{tab:gmm_switch_sensitivity}
\scriptsize
\begin{tabularx}{\columnwidth}{@{}llcX@{}}
\toprule
Factor & Setting & GMM-active layers & Observation \\
\midrule
Depth & 4 hidden layers & $2/8$ & GMM activates on a subset of layers \\
Depth & 6 hidden layers & $3/12$ & Conditional routing remains active \\
Depth & 8 hidden layers & $6/16$ & Behavior persists at greater tested depth \\
$\Delta\mathrm{BIC}$ & 5 / 10 / 20 & $57/228$ each & Identical decisions over the tested range \\
$\Delta\mathrm{BIC}$ & 40 & $22/228$ & Agreement with the default decreases to 0.8465 \\
$\pi_{\ell,L}$ cap & 0.35 / 0.50 / 0.65 & $6/57/172$ of 228 & \revision{Low-mean-component cap is materially sensitive} \\
\bottomrule
\end{tabularx}
\end{table}

The depth audit comprises 38 runs and 228 layer records across the Cross-PDE and boundary-layer settings. With the \revision{low-mean-component} cap fixed at 0.50, $\Delta\mathrm{BIC}$ thresholds from 5 to 20 produce identical branch decisions, whereas a threshold of 40 activates fewer GMM branches. By contrast, changing the \revision{low-mean-component} cap substantially changes the intervention frequency. A paired six-repetition performance check at caps 0.35, 0.50, and 0.65 further shows that the larger cap can reduce parameter error while slightly increasing field error. Thus, the GMM branch provides conditional, distribution-driven grouping without imposing a fixed score cutoff or a fixed intervention ratio, but the switching constants are treated as empirical criteria over the tested configurations rather than task-independent optima.
\end{revisionblock}

\subsubsection{Layer protection ablation}

\begin{table}[htbp]
\centering
\caption{\revision{Layer-protection ablation.}}
\label{tab:layer_ablation}
\small
\resizebox{\columnwidth}{!}{%
\begin{tabular}{lcccccc}
\toprule
Method & $n$ & $L_2$ (\%) & Param (\%) & $\alpha$ (\%) & Avg. selective decay & Wilcoxon \\
\midrule
Depth-aware ($\beta_h\!=\!0.55$) & 10 & $0.074\pm0.015$ & $3.782\pm3.068$ & $11.312\pm9.205$ & $0.044\pm0.011$ & \revision{Reference} \\
No protection ($\beta_h\!=\!0$) & 10 & $0.092\pm0.023$ & $7.478\pm2.869$ & $22.395\pm8.592$ & $0.101\pm0.040$ & $p\!=\!0.002$ \\
\bottomrule
\end{tabular}}
\end{table}

TGSR-PINN controls the selective soft decay intensity at different depths through the layer protection coefficient $\beta_h$. Intuitively, shallow layers are more likely to contain input encoding and low-order spatial structures and thus receive stronger protection; deeper layers are more likely to carry source-task-related local patterns and therefore allow more thorough correction. The overall intensity matching control first computes layer protection coefficients under the default $\beta_h=0.55$, then takes a neuron-count-weighted average to obtain a uniform protection coefficient applied to all hidden layers with the same weak-adaptation signal scaling intensity.

Table~\ref{tab:layer_ablation} shows that after removing layer protection, the relative $L_2$ field error increases from 0.074\% to 0.092\%, the average parameter error from 3.782\% to 7.478\%, and the $\alpha$ error from 11.312\% to 22.395\%. Paired statistics further show that the no-protection version produces higher average parameter error on all 10/10 paired repetitions, with a median paired increase of $+3.437$ percentage points and a two-sided Wilcoxon $p=0.002$. This result shows that, on the smooth advection--diffusion task, removing shallow protection increases the average selective-decay intensity from 0.044 to 0.101 and degrades parameter recovery.

\begin{revisionblock}
The no-protection and overall-intensity-matched controls separate the effects of intervention strength and depth-wise allocation. The no-protection result supports the importance of controlling total selective-decay intensity, while the intensity-matched control evaluates how redistributing the same overall strength across depth affects the correction. Together, these controls support shallow protection as a stability mechanism and depth-aware allocation as a task-dependent refinement. The protection strength can be adapted for source--target pairs with different representation mismatch.
\end{revisionblock}

\subsubsection{Selective soft decay mapping form ablation}

\begin{table}[htbp]
\centering
\caption{\revision{Selective soft-decay mapping-form ablation.}}
\label{tab:map_ablation}
\small
\resizebox{\columnwidth}{!}{%
\begin{tabular}{lccccc}
\toprule
Mapping & $n$ & $L_2$ (\%) & Param (\%) & $\alpha$ (\%) & Wilcoxon \\
\midrule
Linear          & 10 & $0.095\pm0.017$ & $7.051\pm3.795$ & $21.106\pm11.380$ & $p\!=\!0.002$ \\
\textbf{Piecewise cubic (default)} & 10 & $\mathbf{0.074\pm0.015}$ & $\mathbf{3.782\pm3.068}$ & $\mathbf{11.312\pm9.205}$ & \revision{Reference} \\
Sigmoid         & 10 & $0.092\pm0.021$ & $7.361\pm3.358$ & $22.040\pm10.077$ & $p\!=\!0.004$ \\
Hard threshold  & 10 & $0.093\pm0.020$ & $6.430\pm3.283$ & $19.249\pm9.852$ & $p\!=\!0.027$ \\
\bottomrule
\end{tabular}}
\end{table}

Eq.~(\ref{eq:decay_map}) uses a piecewise cubic function to map weak-adaptation signals to continuous scaling coefficients. Table~\ref{tab:map_ablation} shows that under this 10-repetition paired setting, the default piecewise cubic mapping achieves the lowest field error, average parameter error, and $\alpha$ error. Relative to the default mapping, linear, sigmoid, and hard threshold mappings increase average parameter error by 3.269, 3.579, and 2.648 percentage points, respectively, and $\alpha$ error by 9.794, 10.728, and 7.937 percentage points, respectively, with two-sided Wilcoxon $p=0.002$, $0.004$, and $0.027$. The relatively small field error differences, compared with the pronounced parameter error differences, suggest that the selective soft decay mapping form has a stronger effect on parameter recovery than on field reconstruction accuracy in this experiment.

\begin{revisionblock}
We further perform a one-factor-at-a-time sensitivity sweep on the high-P\'{e}clet boundary-layer benchmark over six random-seed repetitions. We vary $\gamma_m\in\{0.75,0.85,0.95\}$, $\gamma_{\min}\in\{0.25,0.40,0.55\}$, and mapping exponent $\kappa\in\{1,2,3,4,5\}$. Table~\ref{tab:decay_hyperparameter_sweep} reports the numerical results, and Figure~\ref{fig:decay_hyperparameter_sweep} shows the corresponding error trends. At the default $(\gamma_m,\gamma_{\min},\kappa)=(0.85,0.40,3)$, the mean parameter and field errors are 0.3036\% and 0.0737\%, respectively. The neighboring value $\gamma_{\min}=0.55$ gives a slightly lower mean parameter error (0.2921\%) but a higher mean field error (0.0768\%). We therefore report the tested neighborhood as an empirical range, not as a task-independent optimum.

\begin{table}[tbp]
\centering
\caption{Sensitivity of the selective soft-decay mapping parameters.}
\label{tab:decay_hyperparameter_sweep}
\scriptsize
\setlength{\tabcolsep}{3pt}
\begin{tabular}{@{}lccc@{}}
\toprule
Parameter & Value & Parameter error (\%) & Field error (\%) \\
\midrule
$\gamma_m$ & 0.75 & $0.4512\pm0.2492$ & $0.0811\pm0.0248$ \\
            & \textbf{0.85} & $\mathbf{0.3036\pm0.3279}$ & $\mathbf{0.0737\pm0.0255}$ \\
            & 0.95 & $0.3437\pm0.3534$ & $0.0813\pm0.0246$ \\
\midrule
$\gamma_{\min}$ & 0.25 & $0.4383\pm0.3317$ & $0.0764\pm0.0300$ \\
                 & \textbf{0.40} & $\mathbf{0.3036\pm0.3279}$ & $\mathbf{0.0737\pm0.0255}$ \\
                 & 0.55 & $0.2921\pm0.1995$ & $0.0768\pm0.0248$ \\
\midrule
$\kappa$ & 1 & $0.5239\pm0.6208$ & $0.1007\pm0.0340$ \\
    & 2 & $0.4848\pm0.3088$ & $0.0809\pm0.0286$ \\
    & \textbf{3} & $\mathbf{0.3036\pm0.3279}$ & $\mathbf{0.0737\pm0.0255}$ \\
    & 4 & $0.4858\pm0.2239$ & $0.0786\pm0.0304$ \\
    & 5 & $0.3820\pm0.1469$ & $0.0754\pm0.0236$ \\
\bottomrule
\end{tabular}
\end{table}

\begin{figure}[tbp]
  \centering
  \includegraphics[width=.62\columnwidth]{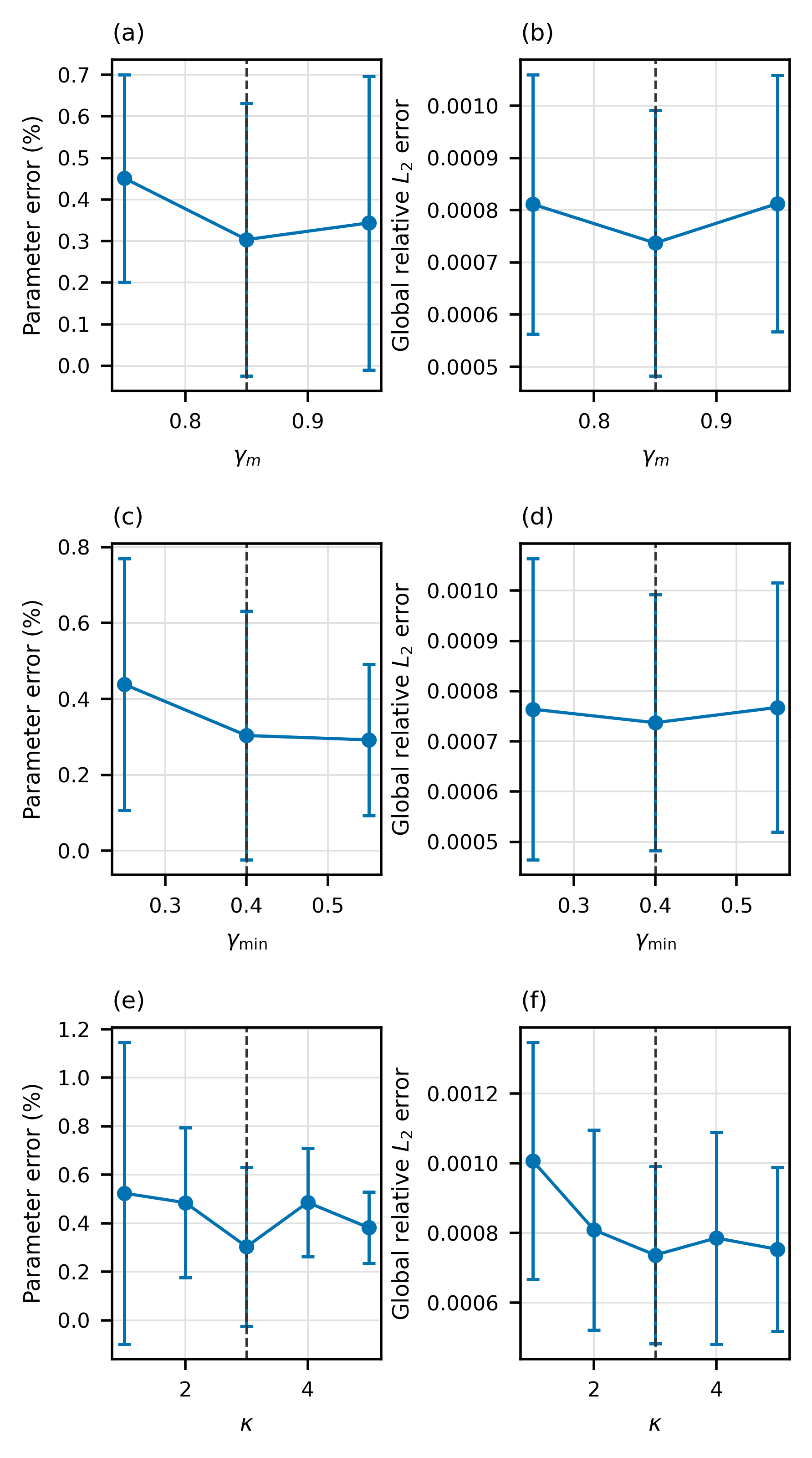}
  \caption{\revision{One-factor sensitivity of the selective soft-decay mapping on the high-P\'{e}clet boundary-layer benchmark. Points and error bars show the mean and standard deviation. Dashed lines identify the default $(\gamma_m,\gamma_{\min},\kappa)=(0.85,0.40,3)$.}}
  \label{fig:decay_hyperparameter_sweep}
\end{figure}
\end{revisionblock}

\begin{revisionblock}
The ablations examine four parts of the correction pipeline. Random Soft Decay and Score-Shuffled TGSR test target-informed selection and score--neuron correspondence. The no-protection variant tests the allocation of correction intensity, and the mapping comparison tests how weak-adaptation signals are converted into bounded scaling factors. The GMM/rank study separately examines layer-wise grouping and fallback routing. Across these controls, the lower parameter errors of TGSR-PINN support the use of the target score to choose where and how strongly the transferred representation is corrected.
\end{revisionblock}

\section{Discussion}
\label{sec:discussion}

\subsection{Joint Analysis of Field Error and Parameter Error}
\label{subsec:field_param}

In PINN inverse problems, physical field errors and physical parameter errors are not always synchronized. The network may compensate for parameter bias by adjusting the field representation, achieving low field prediction errors while still failing to recover the correct target physical parameters. Therefore, this paper uses both relative $L_2$ field error and parameter error as the basis for evaluating inverse transfer effectiveness, rather than judging transfer success solely by field reconstruction accuracy.

\begin{revisionblock}
The local compensation relation introduced in Eqs.~(\ref{eq:compensation_linearization})--(\ref{eq:compensation_direction}) motivates evaluating field reconstruction and parameter recovery separately.

Across the three experiments, TGSR-PINN has the lowest mean parameter error while maintaining low field error. The high-P\'{e}clet benchmark provides a direct example of why both metrics are needed: TL-gPINN has a slightly lower mean field error than TGSR-PINN (0.0860\% versus 0.0946\%) but a higher mean parameter error (0.2510\% versus 0.1260\%). The Cross-PDE and noisy reaction--diffusion results show the same need to evaluate reconstruction and parameter identification separately.

Equations~(\ref{eq:negative_transfer}) and~(\ref{eq:masked_negative_transfer}) express field--parameter decoupling as matched-repetition events. A masked event occurs when transfer preserves or improves field prediction while degrading parameter identification relative to standard PINNs. Table~\ref{tab:negative_transfer_frequency} shows 6--8 parameter-negative-transfer events in ten Full repetitions across the three tasks, compared with 0--3 for TGSR-PINN. The event counts connect the joint field--parameter evaluation with the target-side correction applied before main optimization.

Figure~\ref{fig:field_parameter_relation} plots the two evaluation dimensions jointly for every method and paired repetition. The full-range panel retains unstable runs, and the two detail panels resolve the converged low-error regimes without deleting observations. The scatter shows that methods occupying the same low-field-error region can exhibit materially different parameter errors, supporting the joint reporting of field and parameter metrics rather than using field reconstruction as a proxy for inverse-parameter accuracy.
\end{revisionblock}

\subsection{Analysis of Selective Soft Decay}
\label{subsec:soft_decay_analysis}

\begin{revisionblock}
The following ablations examine the contribution and scope of the correction stages defined in Section~\ref{subsec:tgsr}.
\end{revisionblock}

\begin{revisionblock}
The mechanism ablations on the smooth advection--diffusion task distinguish selective soft decay from short adaptation, random resetting, arbitrary soft perturbation, and score shuffling. In both the early diagnostic and the paired counterfactual study, TGSR-PINN gives lower parameter error than these controls. The layer-protection and mapping ablations further show that the allocation and magnitude of the correction affect parameter recovery.
\end{revisionblock}

\begin{revisionblock}
Selective soft decay is also distinct from ordinary $L_2$ regularization and L2-SP~\cite{li2018l2sp}. Ordinary $L_2$ weight decay adds a global penalty such as $\mu\|\theta\|_2^2$ and continuously shrinks all network parameters toward zero. L2-SP instead uses $\mu\|\theta-\theta_S^*\|_2^2$ to pull all transferred parameters toward the source state throughout target optimization. TGSR-PINN performs one target-conditioned, neuron-selective state correction after short adaptation. The corrected parameters then remain trainable under the original target loss, without a persistent zero or source-state penalty. The L2-SP comparison isolates the difference between persistent global regularization and target-conditioned selective correction.
\end{revisionblock}

\begin{revisionblock}
PINN inverse problems may reach compensatory states with acceptable field predictions but inaccurate physical parameters. Network representations and physical parameters are optimized jointly, so transferred features may steer training toward such states. Selective soft decay changes the pre-activation magnitude of low-scoring neurons before main training. This bounded state correction modifies the optimization starting point and can reduce compensation induced by mismatched transferred representations.
\end{revisionblock}

\begin{figure}[tbp]
  \centering
  \includegraphics[width=.94\columnwidth]{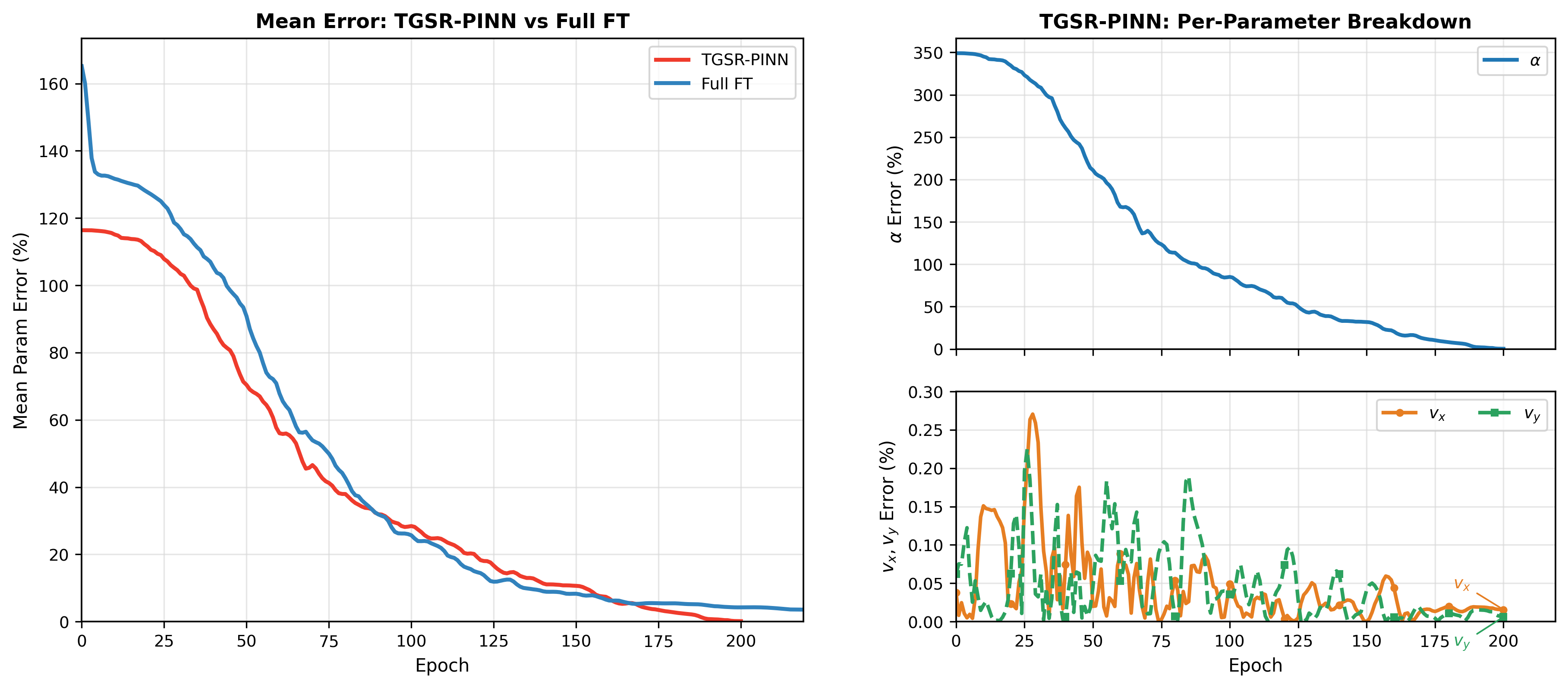}
  \caption{Parameter error convergence curves on the \revision{smooth manufactured-solution advection--diffusion task}.}
  \label{fig:pe_param_post}
\end{figure}

\begin{figure}[tbp]
  \centering
  \includegraphics[width=.94\columnwidth]{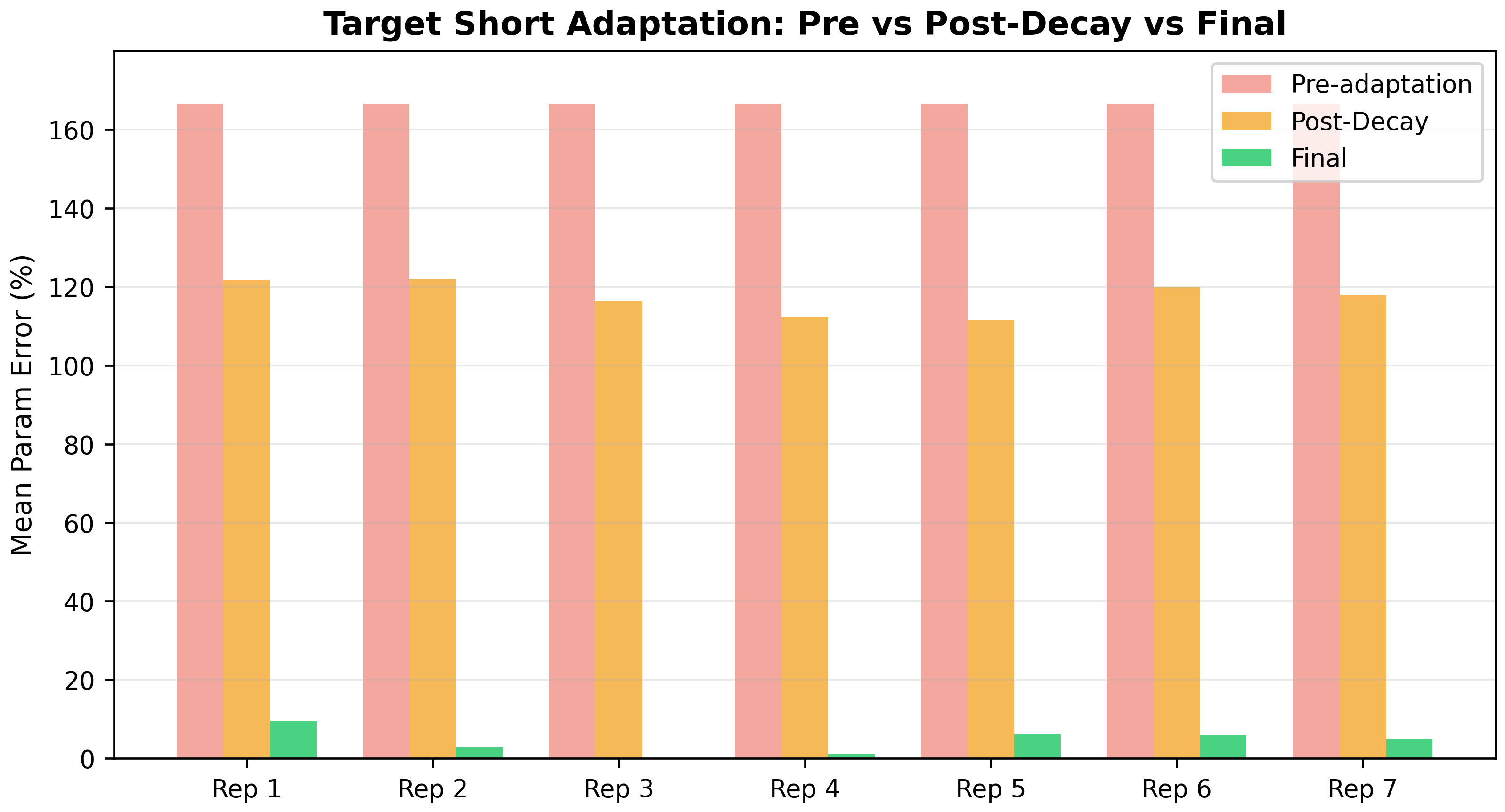}
  \caption{Parameter error changes before and after target short adaptation and selective soft decay on the \revision{smooth manufactured-solution advection--diffusion task}.}
  \label{fig:param_change}
\end{figure}

Figure~\ref{fig:pe_param_post} shows the parameter-error convergence curves during target training, whereas Figure~\ref{fig:param_change} compares the parameter errors before target short adaptation, after selective soft decay, and at the final training stage. \revision{The two figures show that selective soft decay changes the subsequent parameter-recovery trajectory rather than only the field error.}

\begin{figure}[tbp]
  \centering
  \includegraphics[width=.94\columnwidth]{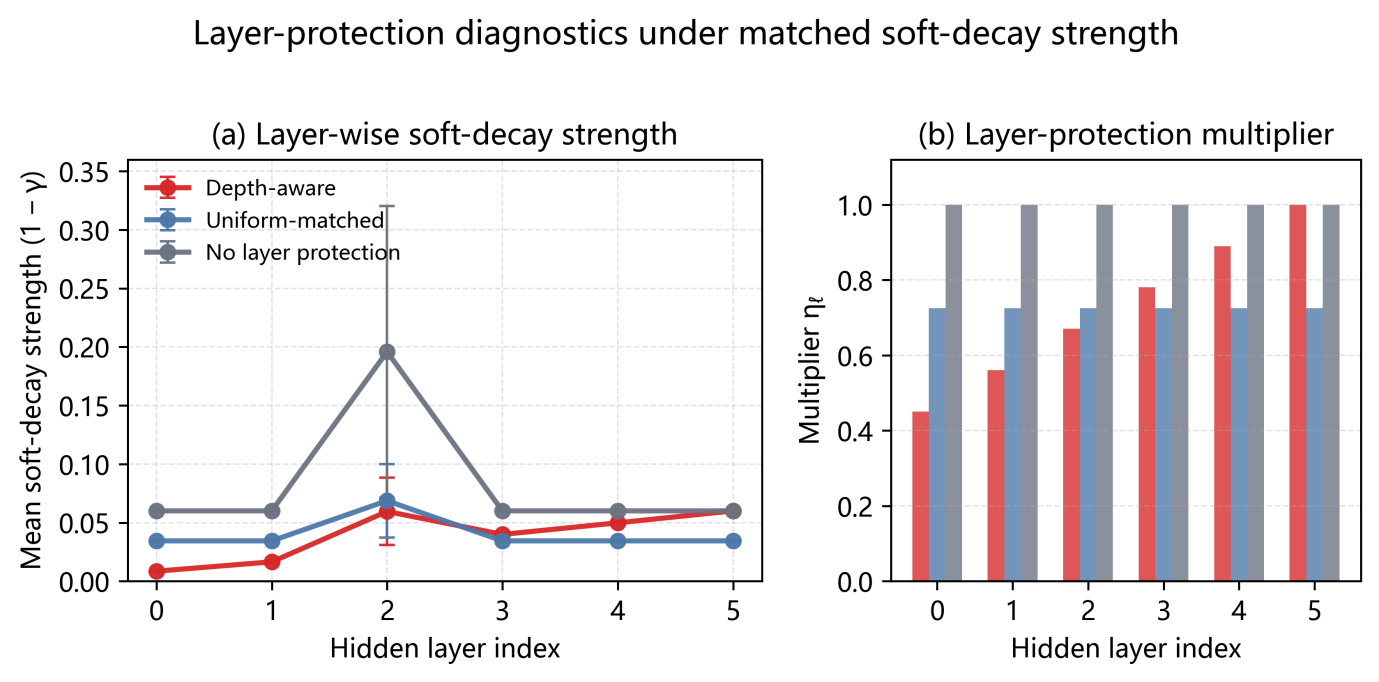}
\caption{Layer-wise selective soft decay diagnosis for layer protection and overall intensity matching.}
  \label{fig:layer_diag}
\end{figure}

\begin{revisionblock}
Layer protection is motivated by a depth-dependent transfer hypothesis. Shallow representations may retain general input encoding and low-order spatial structure, whereas deeper representations may carry more source-specific local patterns. TGSR-PINN therefore applies weaker correction to shallow layers and allows stronger correction at greater depth. Figure~\ref{fig:layer_diag}(a) compares the layer-wise decay intensity under depth-aware protection, overall-intensity matching, and no protection. Figure~\ref{fig:layer_diag}(b) shows the corresponding protection coefficients. Table~\ref{tab:layer_ablation} shows that removing protection increases the total correction intensity and parameter error. The intensity-matched control separates the effect of total intervention strength from its depth-wise allocation.
\end{revisionblock}

\begin{revisionblock}
TGSR-PINN and traditional pruning have different objectives. Traditional pruning typically targets model compression or faster inference. TGSR-PINN instead adapts transferred representations for a target inverse problem. Selective soft decay continuously reduces the influence of low-scoring neurons without deleting or resetting them. The weak-adaptation signal controls this correction using the target loss and target-sample responses, while all neurons remain trainable.
\end{revisionblock}

\subsection{Computational Overhead and Applicability Boundaries}
\label{subsec:cost}

\begin{table}[htbp]
\centering
\caption{TGSR-PINN additional computational overhead breakdown.}
\label{tab:cost}
\small
\begin{tabularx}{\columnwidth}{@{}XccX@{}}
\toprule
Component & Time (s) & Memory (MB) & Notes \\
\midrule
Target short adaptation & $\sim$45  & $\sim$234 & 20\% of main training \\
Neuron scoring          & $\sim$3   & $\sim$470 & Forward+backward \\
GMM + selective soft decay & $\sim$2   & $<$1      & 1D distribution \\
\textbf{Total}          & $\sim$50  & $\sim$470 & 13--19\% of \revision{Full} \\
\bottomrule
\end{tabularx}
\end{table}

TGSR-PINN introduces three additional steps on top of \revision{Full}: target short adaptation, neuron target scoring, and selective soft decay. Table~\ref{tab:cost} provides the time and memory overhead for each step. \begin{revisionblock}The measurements use the time-dependent smooth manufactured-solution advection--diffusion diagnostic with 10000 PDE points, 2500 boundary points, 2500 initial points, and 1000 data points. The workstation runs Windows 11 with an AMD Ryzen 5 9600X CPU and an NVIDIA GeForce RTX 5070 GPU (12 GB). The software environment uses Python 3.13.7, PyTorch 2.11.0.dev20260127+cu128, and CUDA 12.8. The reported times cover training and the additional TGSR-PINN steps. Offline data generation, plotting, dense-grid evaluation, and result aggregation are excluded because they are identical across methods.\end{revisionblock}

\begin{revisionblock}
All methods use the same batch sizes and LBFGS implementation. Each outer epoch uses \texttt{max\_iter=20}, \texttt{history\_size=100}, and a fixed batch for stable closure evaluation. Early stopping depends only on training-loss improvement. It is triggered after 10 rounds without sufficient improvement or when the adjacent loss change falls below $10^{-8}$. Test field error and true parameter error are not used for stopping. Because stopping times vary across repetitions, Table~\ref{tab:cost} reports empirical runtime rather than hardware-independent complexity. On the smooth manufactured-solution task, TGSR-PINN requires 266 s, compared with 235 s for Full, corresponding to approximately 13\% overhead. The Cross-PDE overhead is 19\% because its main training stage is shorter. These measurements characterize the cost of target-side correction; they do not imply general training acceleration.
\end{revisionblock}

GMM fitting performs two-component EM estimation on the 1D target-score distribution of 100 neurons per layer. Its complexity does not depend directly on whether the physical problem is one-, two-, or three-dimensional. \begin{revisionblock}At the tested network scale, GMM fitting and selective soft decay take much less time than target short adaptation and neuron scoring. Target short adaptation is therefore the main source of additional cost. Its cost can be reduced by adjusting the adaptation budget $E_{ad}$. GMM fitting remains a low-overhead step that supplies layer-wise distributional information for weak-adaptation signal estimation.\end{revisionblock}

\begin{revisionblock}
The empirical scope of TGSR-PINN is inverse transfer in which source and target tasks retain partially reusable representations while direct fine-tuning can bias physical-parameter recovery. Its target-side diagnosis requires short adaptation to produce informative loss gradients and pre-activation responses. Consequently, extremely sparse observations, severely misaligned parameter initialization, unstable short adaptation, or substantial model-form error can contaminate the neuron scores and weaken the subsequent correction. The tested boundary-layer, Cross-PDE, and noisy reaction--diffusion problems span distinct forms of source--target mismatch, but three discrete task families are insufficient to estimate a universal threshold in P\'{e}clet-number gap, observation noise, or PDE-family discrepancy. The present results therefore define an empirical applicability range rather than a task-independent discrepancy boundary. Likewise, the depth and GMM-threshold diagnostics support the tested configurations but do not justify extrapolation to arbitrary network depths or score distributions. Validation under unknown boundary conditions, biased measurements, model discrepancy, and real engineering data remains necessary.
\end{revisionblock}

\section{Conclusion}
\label{sec:conclusion}

\begin{revisionblock}
This paper proposes TGSR-PINN, a target-evidence-driven neuron-level representation adaptation method for PINN inverse transfer learning. Source physical parameters are not inherited. After target short adaptation, the method combines Taylor sensitivity and pre-activation variance to score transferred neurons, estimates continuous weak-adaptation signals using a conditional GMM branch with rank fallback, and applies bounded selective soft decay while keeping every neuron trainable.

The evaluation covers a zero-source high-P\'{e}clet inflow--outflow benchmark with nonzero Dirichlet boundary data, Allen--Cahn to Burgers cross-PDE transfer, and 5\%-noise reaction--diffusion inversion. Across the paired-repetition cohorts, TGSR-PINN achieves the lowest mean parameter error among the compared methods in all three tasks and the lowest mean field error in the latter two. On the high-P\'{e}clet boundary-layer benchmark, it reduces the mean parameter error to 0.1260\% while maintaining a mean field error of 0.0946\%. Paired wins and task-specific Wilcoxon tests further support the observed parameter-recovery advantage in several baseline comparisons.

Paired counterfactual ablations indicate that target-informed neuron selection and score--neuron correspondence contribute to parameter recovery relative to random soft decay and score-shuffled variants. Layer protection and the bounded piecewise-cubic mapping may also help limit over-correction in the tested settings. The GMM/rank study shows that the estimator conditionally uses layer-specific mixture grouping when supported by the score distribution and otherwise retains a bounded rank-based signal, thereby avoiding fixed cutoffs and fixed intervention ratios. Overall, TGSR-PINN converts coarse transferred-weight reuse into a target-evidence-driven diagnosis and recoverable correction process. \revision{Future theoretical work will examine convergence, local identifiability, and optimization under bounded soft decay. Method development will consider adaptive intervention bounds, identifiability-aware scoring, low-rank and multi-source transfer, and neural-operator integration. Engineering validation will address nondestructive testing, multiphysics inverse problems, model discrepancy, and biased measurements.}
\end{revisionblock}


\section*{Code Availability}
The source code, configuration files, and scripts for reproducing the experiments in this paper are publicly available at \url{https://github.com/jooycelee/TGSR-Pinns}. The repository includes the TGSR-PINN implementation, benchmark task configurations, and post-processing utilities used to generate the reported tables and figures.


\begin{revisionblock}
\section*{Declaration of Generative AI and AI-assisted Technologies in the Writing Process}
During the preparation of this work, the authors used OpenAI ChatGPT to improve the language and readability of the manuscript. After using this tool, the authors reviewed and edited the content as needed and take full responsibility for the content of the publication.
\end{revisionblock}


\section*{Acknowledgements}
This work was supported by the Natural Science Foundation of Fujian Province, China (Grant Nos. 2026J008225 and 2026J008226), and the Foundation of Fujian Provincial Department of Education (Grant Nos. JAT251084 and JAT251087). The authors also thank the anonymous reviewers for their constructive comments that helped improve this paper.


\end{document}